\documentclass[a4paper,fleqn]{cas-dc}

\usepackage[authoryear]{natbib}
\usepackage{soul}
\usepackage{pifont}
\usepackage{algorithm}
\usepackage{algorithmic}
\usepackage{subcaption}

\usepackage[normalem]{ulem}
\usepackage{soul}
\sethlcolor{blue!15}

\begin{document}
\let\WriteBookmarks\relax
\def\floatpagepagefraction{1}
\def\textpagefraction{.001}

\shorttitle{EV9V: An Echocardiographic Video Benchmark}

\shortauthors{}

\title[mode=title]{EV9V: An Echocardiographic Video Benchmark with Uncertainty Guided View Classification}

\author[1]{Bo Gou}
\fnmark[$\ast$]
\author[1]{Jicheng Zhang}
\fnmark[$\ast$]
\author[2]{Jianlong Xiong}
\author[2]{Tao He}
\author[1]{Hai Wu}
\author[1]{Bentian Liu}
\author[1]{Yijiao Wang}
\author[1]{Yu Zhang}
\author[3]{Yujia Yang}
\author[4]{Yun Dai}
\author[3]{Wei Liu}
\author[1]{Jian Liu}
\fnmark[$\dagger$]
\ead{lj190090998@cmc.edu.cn}
\author[1]{Jie Wang}
\fnmark[$\dagger$]
\ead{wj0815@cmc.edu.cn}

\address[1]{Department of Ultrasound, Clinical Medical College and the First Affiliated Hospital of Chengdu Medical College, Chengdu, China} 
\address[2]{College of Computer Science, Sichuan University, Chengdu, China} 
\address[3]{West China Hospital of Sichuan University, Chengdu, China} 
\address[4]{Cancer Hospital, Chinese Academy of Medical Sciences and Peking Union Medical College, Beijing, China} 
\nonumnote{$^\ast$ They are co-first authors and contribute equally to this study.}
\nonumnote{$^\dagger$ Corresponding author.}

\begin{abstract}
Automated classification of standard echocardiographic views is crucial for efficient clinical workflow but faces three main challenges. First, publicly available datasets are scarce and limited in scale and view coverage. Second, the performance of some modern video-level architectures for echocardiographic view classification remains underexplored. Third, some view categories exhibit highly similar spatial appearances, making single-frame features insufficient for discrimination, while heterogeneous frame quality complicates robust temporal information fusion. To address these challenges, we release the Echocardiographic Videos of Nine Views (EV9V) dataset, comprising 5,138 videos, 910,579 frames, and 9 standard views, which is, to the best of our knowledge, the largest openly released echocardiography video dataset. Furthermore, we propose a Spatio-Temporal Fusion Model (STFM), an efficient dual-stream framework that jointly captures spatial anatomical structures and temporal cardiac dynamics. The STFM leverages uncertainty-aware learning to preferentially sample representative video segments during training and evidence-based fusion during inference, improving robustness to variations in frame quality across echocardiographic videos. Extensive experiments demonstrate that STFM achieves competitive performance and effectiveness compared with diverse video classification architectures, including two-dimensional and three-dimensional convolutional neural networks, recurrent neural networks, and Transformers. The code is released at \url{https://huggingface.co/datasets/bgx666/EV9V} and the EV9V dataset is available at \url{https://github.com/bgx666/stfm}.
\end{abstract}


\maketitle

\section{Introduction}
Echocardiography is one of the most widely used imaging modalities in cardiovascular diagnosis due to its real-time, noninvasive, and cost-effective nature. Clinical assessment typically requires the integration of multiple standard planes, each corresponding to specific anatomical structures and diagnostic priorities; therefore, the accurate recognition of these views serves as a fundamental prerequisite for the reliable evaluation of cardiac structure and function. Clinically, this process relies heavily on the expertise of ultrasound physicians, requiring substantial training and experience to ensure diagnostic accuracy. However, clinical practice often faces challenges such as increasing patient volumes, limited availability of experienced sonographers, and the inherent operator dependence of ultrasound imaging. Furthermore, the development of advanced diagnostic models necessitates the precise and rapid delivery of categorized input views to ensure reliable analysis \citep{barrios2026multiview}. Automated recognition of standard echocardiographic planes offers a promising solution to these challenges. By enabling real-time, objective identification of standard views, artificial intelligence (AI)–driven systems can assist clinicians in rapidly obtaining diagnostically relevant images, thereby improving workflow efficiency and reducing inter-observer variability. Several routine steps rest on this classification in practice: an examination can be checked, while the patient is still on the table, for the standard planes it should contain but does not~\citep{shi2026clinically}; the measurements a cardiologist compares between follow-up visits, such as cavity dimensions and ejection fraction, are defined on particular views and lose their meaning when a similar view is mistaken for one of them~\citep{lang2015recommendations}. 

Automated classification of standard views in echocardiographic data is widely studied in early years. However, most of them focused on image-level evaluation on limited small dataset. \citet{zhou2006image} tackled the classification of standard views as a multiple binary classifier boosting procedure, which used the logitboost to fuse weak classifiers into a strong one. For enhancing the classification accuracy, \citet{khamis2017automatic} proposed a multi-stage classification algorithm, which is composed of a spatio-temporal feature extraction and a discriminative learning dictionary modules. \citet{kusunose2020clinically} constructed a dataset of $17,000$ images, and evaluated this dataset using three general convolutional neural networks (CNNs). \citet{gao2021automated} proposed a three-stage recognition method, where a graph-based image embedding algorithm is used to improve the generalization accuracy.

In recent years, with the development of deep learning, video-level standard view classification in echocardiographic data has been developed. Constructing a model based on deep neural networks and investigating information fusion methods has been developed. The importance of spatio-temporal feature fusion was first well analyzed by \citet{gao2017fused}. They built a two-stream CNN architecture for feature fusion, where one 2D CNN was used for processing a fixed-length video at the spatial steam and another 2D CNN was used for extracting temporal features with the help of Optical Flow. \citet{madani2018fast} built a VGG-based model for 15 standard views. \citet{howard2020improving} discovered that temporal features among echocardiographic videos were important in standard view classification, therefore they evaluated time-distributed networks and two-stream networks on $9,098$ echocardiographic videos. \citet{ostvik2019real} focused on the recognition efficiency for video-level standard view classification. An InceptionV3-based \citep{szegedy2016rethinking} model was built and achieved a $4.4$ ms per frame real-time performance. \citet{naser2024artificial} constructed a huge echocardiographic video dataset, including $10,269$ videos from $909$ patients. They compared the 2D and 3D CNNs using ResNet18, InceptionV3, VGG13, and VGG16 baselines. \citet{mohammadi2024presenting} used transfer learning and built a 3D InceptionV3 model to improve the performance of standard view classification. TTESlowFast \citep{cheng2025deep} used a sampling balance method to extract region of interest followed by a two-stream architecture, where a voting method is used in their inference stage.  

\begin{table*}[!tb]
  \centering
  \caption{Statistics of related echocardiographic video datasets.}
  \setlength{\tabcolsep}{13pt}
\begin{tabular}{lccccccc}
    \toprule
    data sources & years & videos & frames & pixels & categories & availability \\
    \midrule
    Gao et al.\ (\citeyear{gao2017fused}) & 2017  & 432   & 11,232 & 434$\times$636 & 8     & \sethlcolor{red}\hl{\textbf{private}} \\
    Madani et al.\ (\citeyear{madani2018fast}) & 2018  & 267   & 834,267 & 60$\times$80 & 15  & \sethlcolor{gray}\hl{\textbf{on request}} \\ 
    {\O}stvik et al.\ (\citeyear{ostvik2019real}) & 2019  & 7,141 & 496,600 & 128$\times$128 & 7     & \sethlcolor{red}\hl{\textbf{private}} \\
    Howard et al.\ (\citeyear{howard2020improving}) & 2020  & 9,098 & unknown & 299$\times$299 & 14 & \sethlcolor{red}\hl{\textbf{private}} \\
    Azarmehr et al.\ (\citeyear{azarmehr2021neural}) & 2021  & 8,732 & 41,321 & 128$\times$128 & 14 & \sethlcolor{red}\hl{\textbf{private}} \\
    Mohammadi et al.\ (\citeyear{mohammadi2024presenting}) & 2024  & 840   & 36,509 & 480$\times$640 & 7     & \sethlcolor{red}\hl{\textbf{private}} \\
    Naser et al.\ (\citeyear{naser2024artificial}) & 2024  & 11,136 & unknown & 256$\times$256 & 9  & \sethlcolor{gray}\hl{\textbf{on request}} \\
    Cheng et al.\ (\citeyear{cheng2025deep}) & 2025  & 6,013 & 687,493 & 224$\times$224 & 9     & \sethlcolor{gray}\hl{\textbf{on request}} \\
    Elmekki et al.\ (\citeyear{elmekki2025cactusopendatasetframework})&2025 & unknown & 37,736 & 224$\times$224 & 5 &\sethlcolor{green}\hl{\textbf{public}} \\
    \midrule
    EV9V (Ours) & 2025  & 5,138 &\sethlcolor{green}\hl{\textbf{910,579}} & 320$\times$240 & 9     & \sethlcolor{green}\hl{\textbf{public}} \\
    \bottomrule
    \end{tabular} 
  \label{tab:dataset_comp}%
\end{table*}%
Though substantial progress has been made in automated standard-view classification for echocardiographic videos, several important limitations remain.

\ding{182} \textbf{Large-scale publicly accessible datasets remain scarce.}
Table~\ref{tab:dataset_comp} summarizes representative echocardiographic video datasets published between 2017 and 2025. Among them, only a few are available upon reasonable request, while most remain private and inaccessible to the research community. The limited availability of large-scale open datasets hinders reproducible research, fair comparison, and the development of data-hungry deep learning models.

\ding{183} \textbf{Standardized benchmarking of modern video architectures is lacking.}
Previous studies primarily evaluated conventional CNN backbones, including VGG, Inception, ResNet, and DenseNet \citep{gao2017fused, madani2018fast, ostvik2019real, naser2024artificial}. Most of these architectures were proposed before 2020 and therefore do not reflect recent advances in video understanding. Furthermore, previous studies were conducted on different datasets and employed varying view definitions, data partition protocols, and evaluation metrics, making direct comparison difficult and hindering systematic assessment of modern video architectures for echocardiographic view classification.

\ding{184} \textbf{Reliable and efficient spatio-temporal information fusion remains challenging.}
Many standard echocardiographic views exhibit highly similar anatomical appearances in individual frames and can only be reliably distinguished through characteristic temporal motion patterns \citep{howard2020improving}. Existing approaches predominantly rely on computationally expensive 3D CNNs \citep{7410867,carreira2017quo} or optical-flow-based two-stream architectures \citep{simonyan2014two} for spatio-temporal modeling. While 3D CNNs incur substantial computational overhead, optical-flow-based methods require additional motion estimation and are susceptible to ultrasound speckle noise. Moreover, echocardiographic videos frequently contain frames of varying informativeness due to probe motion, respiratory artifacts, and transitional cardiac phases. Existing methods typically aggregate predictions from multiple sampled frames or clips through probability averaging or voting mechanisms. Consequently, overconfident predictions from non-prototypical frames may propagate to the video level and compromise the reliability of the final classification result.

To address these limitations, we construct and publicly release the Echocardiographic Videos of Nine Views (EV9V) dataset and propose a Spatio-Temporal Fusion Model (STFM) for video-level standard-view classification. The main contributions are:

\begin{itemize}
    \item We publicly release the EV9V dataset, comprising 9 standard echocardiographic views, 5,138 videos, and 910,579 frames. To our knowledge, EV9V is the largest publicly accessible echocardiography video dataset, featuring comprehensive view coverage, rigorous three-stage physician quality control, and substantial real-world variability in acquisition duration and imaging conditions.
    
    \item We establish a comprehensive benchmark for video-level echocardiographic view classification on EV9V, systematically evaluating a diverse range of modern video architectures, including 2D CNNs, 2D CNNs with temporal modeling, RNNs, 3D CNNs, and video transformers.

    \item  We propose STFM, an efficient dual-stream CNN--LSTM architecture that jointly models spatial anatomy and temporal cardiac dynamics. Combined with evidential uncertainty modeling and an uncertainty-guided segment selection strategy, the framework mitigates the impact of non-prototypical frames and improves the reliability of video-level echocardiographic view classification.
\end{itemize}

The remainder of this paper is structured as follows. 
Section~\ref{sec:relate_work} reviews prior work on 2D and 3D echocardiographic view classification. 
Section~\ref{sec:EV9V_dataset} details the construction and statistical distribution of the proposed EV9V dataset. 
Section~\ref{sec:Method} presents STFM, our proposed spatiotemporal fusion model for view classification. 
In Section~\ref{sec:baseline_evaluation}, we systematically evaluate a diverse range of models including 2D CNNs, 3D spatiotemporal architectures, and our proposed STFM on EV9V to establish comprehensive benchmarks. 
Section~\ref{sec:ablation} reports ablation studies isolating the key components of STFM. 
Section~\ref{sec:interpretability} offers a visual interpretability analysis of the model's predictions. 
Section~\ref{sec:discussion} interprets the findings and states the limitations of the present study.
Finally, Section~\ref{sec:conclusion} summarizes the main findings.

\section{Related Work}
\label{sec:relate_work}

\subsection{Image Classification Architectures}
The evolution of modern computer vision architectures began with pioneering designs like VGG \citep{simonyan2015very}, which demonstrated that increasing network depth using uniform convolutions significantly enhanced representational power. However, training deeper networks introduced optimization challenges, which were elegantly resolved by ResNet \citep{he2016deep} through residual connections, enabling the effective training of widely adopted models such as ResNet-18 and ResNet-50. Building on this, DenseNet \citep{huang2017densely} introduced dense layer-wise connections to maximize feature reuse and strengthen gradient flow, yielding highly efficient architectures like DenseNet-121 and DenseNet-169. As the focus expanded to deployment on resource-constrained devices, lightweight designs emerged, notably MobileNetV3 \citep{Koonce2021}, which utilized hardware-aware architecture search to create efficient variants including MobileNetV3-Small and Large. To systematically scale networks beyond empirical tuning, EfficientNet \citep{tan2019efficientnet,tan2021efficientnetv2} proposed a compound scaling strategy balancing depth, width, and resolution (e.g., EfficientNet-B7, EfficientNetV2-L), while RegNet \citep{radosavovic2020designing} optimized the network structural design space. Recently, ConvNeXt \citep{liu2022convnet} modernized standard CNNs by incorporating large kernels, inverted bottlenecks, and modern normalizations, proving that pure convolutional models (e.g., ConvNeXt-Tiny, Small, and Base) remain highly competitive against modern hybrid architectures.

Beyond CNNs, Transformer-based self-attention models have profoundly reshaped visual recognition. The Vision Transformer (ViT) \citep{dosovitskiyimage} successfully adapted pure self-attention to global image patches, bypassing convolutional biases and achieving remarkable performance at scale with variants like ViT-L-16. To address the quadratic computational complexity of global attention, Swin-Transformer \citep{liu2021swin} and its enhanced version SwinV2 \citep{liu2022swin} introduced a hierarchical architecture with shifted local windows, establishing themselves as versatile backbones across diverse vision tasks (e.g., Swin-B, SwinV2-B). Furthermore, hybrid designs such as MaxViT \citep{tu2022maxvit} integrated both local and global attention mechanisms alongside convolutional inductive biases, optimizing the trade-off between accuracy and computational efficiency, as demonstrated by models like MaxViT-T.

\subsection{Video Classification Models}
The evolution of video pattern recognition has transitioned from simple spatial aggregation to sophisticated spatiotemporal modeling. Early methodologies primarily relied on 2D CNNs coupled with temporal aggregation. For instance, the pioneering Two-stream architecture \citep{simonyan2014two} processed spatial appearance and temporal motion through separate pathways. To capture longer contexts, TSN \citep{wang2016temporal} introduced sparse temporal sampling over entire videos, while TSM \citep{lin2019tsm} achieved highly efficient temporal modeling via channel shifting. Subsequent advancements, such as TANet \citep{liu2021tam} and TIN \citep{shao2020temporal}, further enriched temporal reasoning through multi-scale features and interleaving mechanisms. Alternatively, RNN-based approaches address temporal dynamics by integrating spatial feature extractors with recurrent architectures, such as LSTM, GRU, and BiLSTM, to sequentially model spatial trajectories over time.

To concurrently capture spatial and temporal dimensions, 3D CNNs became a prominent paradigm. C3D \citep{7410867} demonstrated the effectiveness of 3D kernels for large-scale video classification, and I3D \citep{carreira2017quo} significantly improved performance by inflating 2D pre-trained weights into 3D structures. To mitigate the immense computational overhead of dense 3D convolutions, architectures like R(2+1)D \citep{Tran2017ACL} decomposed 3D kernels into separable 2D spatial and 1D temporal convolutions, while C2D \citep{wang2018non} established strong 2D convolutional baselines. Furthermore, multi-pathway designs were introduced to optimize efficiency and feature extraction; notably, SlowFast \citep{feichtenhofer2019slowfast} and its SlowOnly baseline separated the processing of fine-grained spatial semantics and rapid motion, while TPN \citep{yang2020tpn} constructed temporal pyramids to capture actions occurring at various speeds.

Driven by the superior long-range dependency modeling of attention mechanisms, Transformers have been extensively adapted for video understanding. TimeSformer \citep{bertasius2021space} pioneered a convolution-free approach using divided space-time self-attention. Hierarchical designs like MViT \citep{li2022mvitv2} and the Video Swin Transformer \citep{liu2021video} incorporated inductive biases through localized spatiotemporal windows. Moreover, UniFormer and its successor UniFormerV2 \citep{li2022uniformerv2} successfully merged the local efficiency of convolutions with the global reach of Transformers, setting competitive benchmarks for action recognition.

\subsection{Deep Learning in Echocardiography}
Automated view classification is a fundamental precursor to comprehensive computer-aided interpretation of echocardiograms \citep{madani2018fast}. Early research explored image-based classification using handcrafted features and traditional machine learning, with \citet{zhou2006image} introducing a multiclass boosting procedure for rapid view detection and \citet{khamis2017automatic} employing spatio-temporal cuboid detectors and dictionary learning to distinguish similar apical views.

With the advent of deep learning, 2D-CNNs achieved expert-level accuracy on single-frame classification by recognizing clinically relevant spatial features \citep{madani2018fast, kusunose2020clinically}. Subsequent innovations addressed robustness, edge deployment, and safety: \citet{gao2021automated} used spatial transform networks to handle shape variability, \citet{azarmehr2021neural} employed architecture search for lightweight models.

Despite the success of frame-level analysis, echocardiography is inherently dynamic, and recent work has shifted toward video-level classification. \citet{ostvik2019real} achieved state-of-the-art accuracy by incorporating temporal sequences, while \citet{howard2020improving} demonstrated that time-distributed and two-stream architectures can effectively track anatomical structures. \citet{naser2024artificial} provided a comprehensive evaluation of 2D and 3D CNNs across diverse clinical settings, highlighting trade-offs in predictive value across views. More recently, \citet{cheng2025deep} introduced balanced sampling and augmentation strategies to address class imbalance. More recently, large-scale foundation models have been applied to echocardiography. \citet{kim2025echofm} trained a self-supervised spatio-temporal framework on over 20 million images and demonstrated robust performance across downstream tasks.

The contribution of STFM lies in how its ingredients are coupled rather than in any single one of them. Dual-stream recognition with an LSTM on top is not new in itself; here the temporal stream consumes the pixels of a short sparsely sampled clip and shares its shallow stem with the spatial stream, so that no optical flow, no second backbone and no fusion of separately trained stream scores is required, which matters for ultrasound because flow estimation is unreliable on speckle. The evidential head is likewise not only a replacement for the softmax loss: the Dirichlet parameters are the quantities from which the video-level prediction is assembled, with the evidence of the sampled segments pooled before the uniform prior is applied in Eq.~(\ref{eq:alpha_fusion}), so that a single fused distribution yields the prediction, its uncertainty and the weight of each observation. The same uncertainty also steers the segment selection of Section~\ref{sec:segment_selection}, so one signal serves both the fused decision and the data on which the model is trained, without any extra component at inference. 

\section{EV9V Dataset}
\label{sec:EV9V_dataset}
\subsection{Dataset Construction}
This study was approved by the Ethics Committee of a grade 3 and first-class hospital (Ethics Approval No.  2026CYFYIRB-BA-067). To protect patient privacy, complete anonymization and removal of all protected health information  were performed prior to any data processing and annotation work. The dataset consists of 5,138 clinical echocardiographic examination videos acquired from the hospital during the period from January 2020 to January 2023. All examinations were performed by licensed ultrasound physicians with varying levels of seniority (from residents to chief physicians) in strict accordance with standardized clinical scanning protocols, capturing natural variations in patient anatomical structure and operator proficiency across the dataset.

\textbf{Cohort screening.} The inclusion criteria were as follows: (1) patients aged between 20 and 68 years; (2) echocardiography performed for routine cardiac assessment or screening for suspected heart disease; (3) the scanning process strictly followed standardized echocardiography examination protocols, with video data containing complete standard cardiac views; (4) no significant motion artifacts, poor acoustic window conditions, severe image blurring, or other conditions that impede the identification of cardiac anatomical structures. The exclusion criteria were defined as follows: (1) incomplete visualization of cardiac structures in the video, with key anatomical structures such as the atria, ventricles, and valves unable to be clearly identified; (2) reduced video quality due to inappropriate patient positioning, severe respiratory motion, or non-standard operation, rendering it unsuitable for further analysis; (3) missing or incomplete clinical diagnostic information for the case, which prevents accurate annotation and classification.

\textbf{Annotation protocol and quality control.} A rigorous three-tier physician quality control system was implemented to ensure the accuracy and consistency of annotations. Videos were initially annotated independently in a double-blind manner by multiple licensed attending physicians with clinical experience in echocardiography, using the Anvil video annotation research tool. Any discrepancies or controversies that arose during this stage were reviewed and adjudicated by higher-ranking associate chief physicians or senior experts. Finally, a secondary verification was conducted by senior experts in cardiac ultrasound diagnosis to confirm the consistency and clinical reliability of all labels.
\begin{figure*}[!tb]
    \centering
    \begin{subfigure}{0.32\textwidth}
        \centering
        \includegraphics[width=\linewidth]{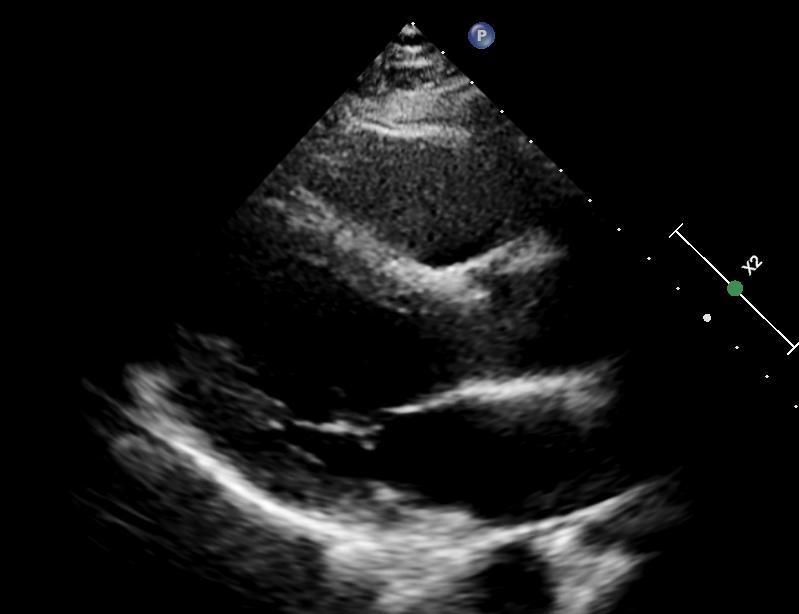}
        \caption{PLHLA}
    \end{subfigure}
    \hfill
    \begin{subfigure}{0.32\textwidth}
        \centering
        \includegraphics[width=\linewidth]{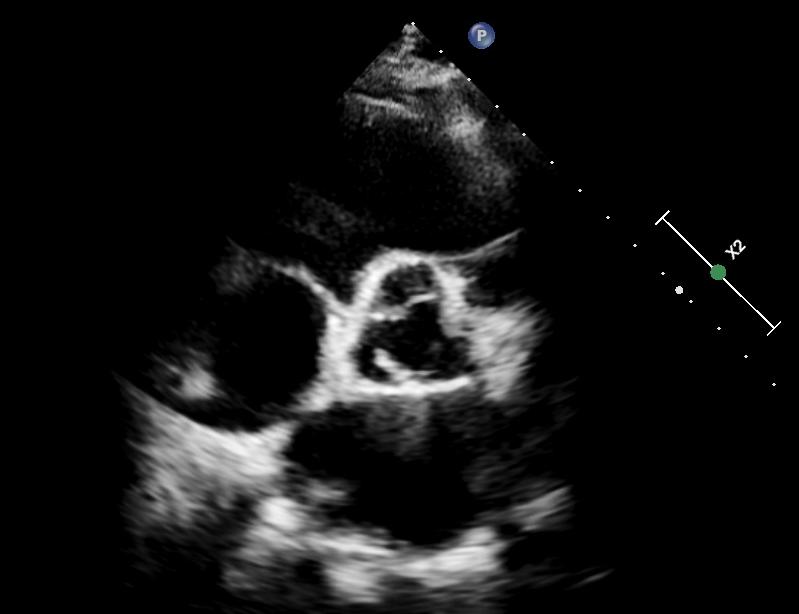}
        \caption{PASA}
    \end{subfigure}
    \hfill
    \begin{subfigure}{0.32\textwidth}
        \centering
        \includegraphics[width=\linewidth]{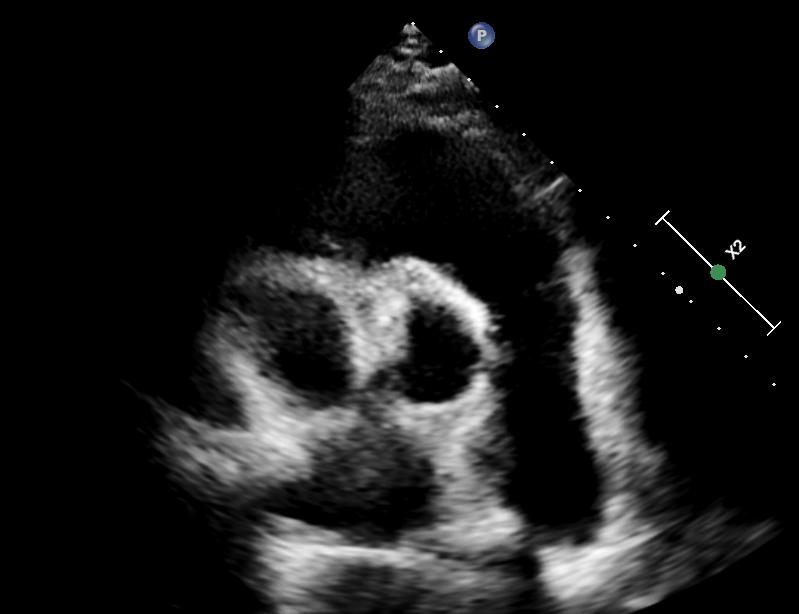}
        \caption{PMPALA}
    \end{subfigure}
    
    \begin{subfigure}{0.32\textwidth}
        \centering
        \includegraphics[width=\linewidth]{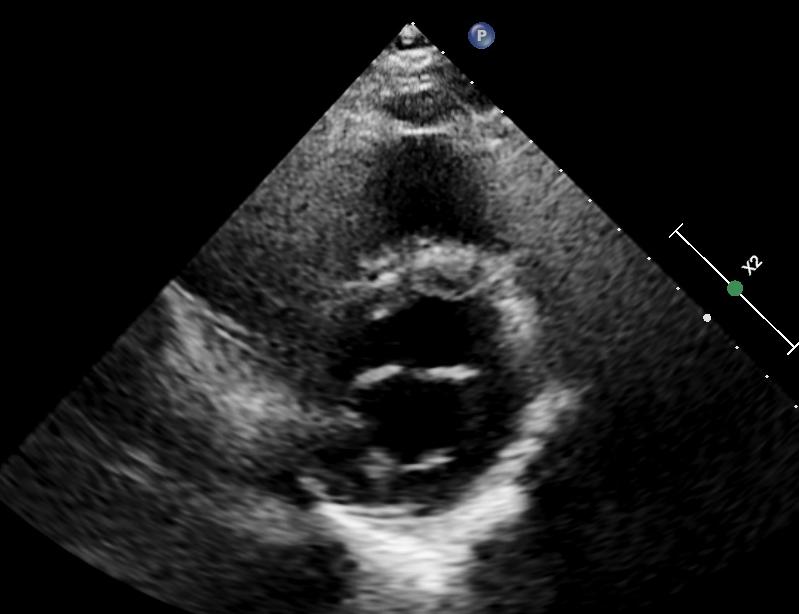}
        \caption{PMVLSA}
    \end{subfigure}
    \hfill
    \begin{subfigure}{0.32\textwidth}
        \centering
        \includegraphics[width=\linewidth]{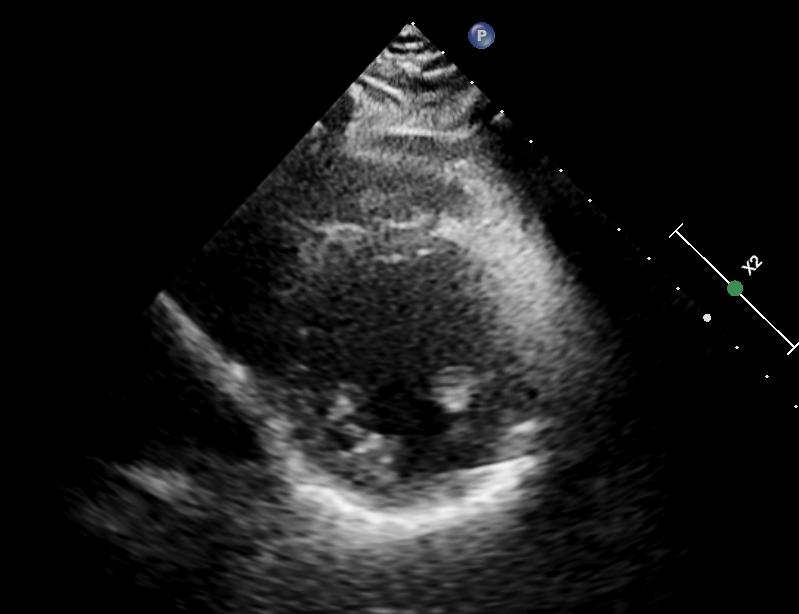}
        \caption{PPMLSA}
    \end{subfigure}
    \hfill
    \begin{subfigure}{0.32\textwidth}
        \centering
        \includegraphics[width=\linewidth]{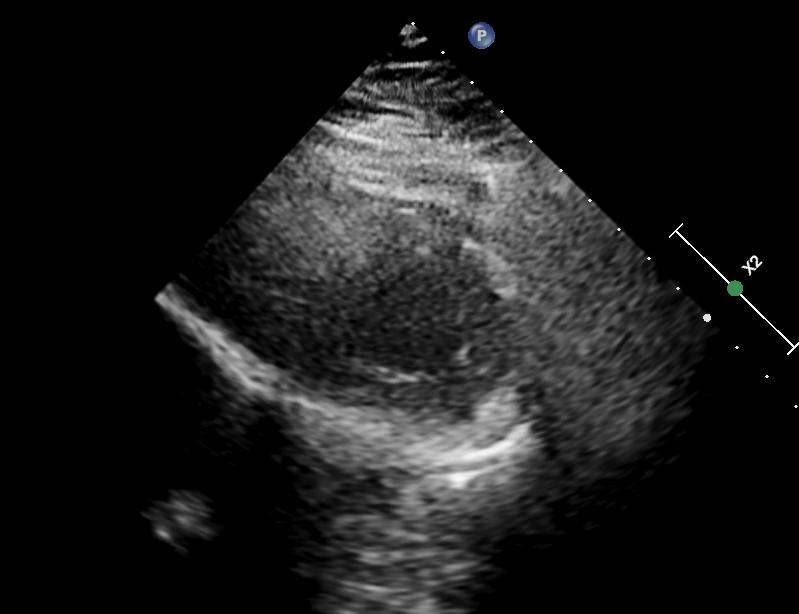}
        \caption{PMASA}
    \end{subfigure}
    
    \begin{subfigure}{0.32\textwidth}
        \centering
        \includegraphics[width=\linewidth]{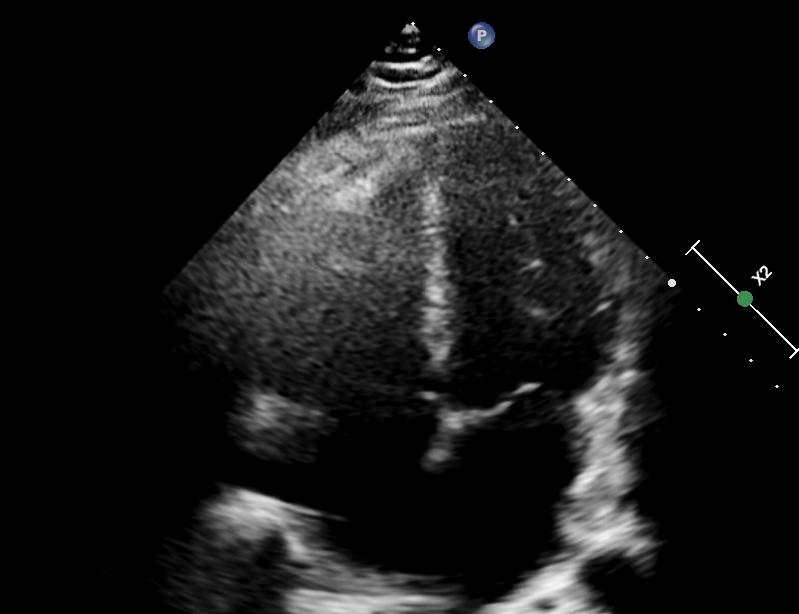}
        \caption{A4C}
    \end{subfigure}
    \hfill
    \begin{subfigure}{0.32\textwidth}
        \centering
        \includegraphics[width=\linewidth]{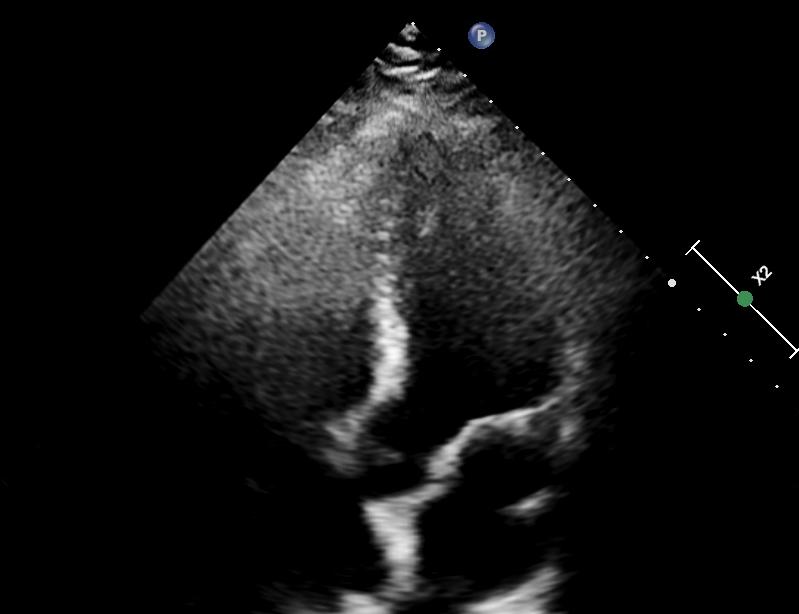}
        \caption{A5C}
    \end{subfigure}
    \hfill
    \begin{subfigure}{0.32\textwidth}
        \centering
        \includegraphics[width=\linewidth]{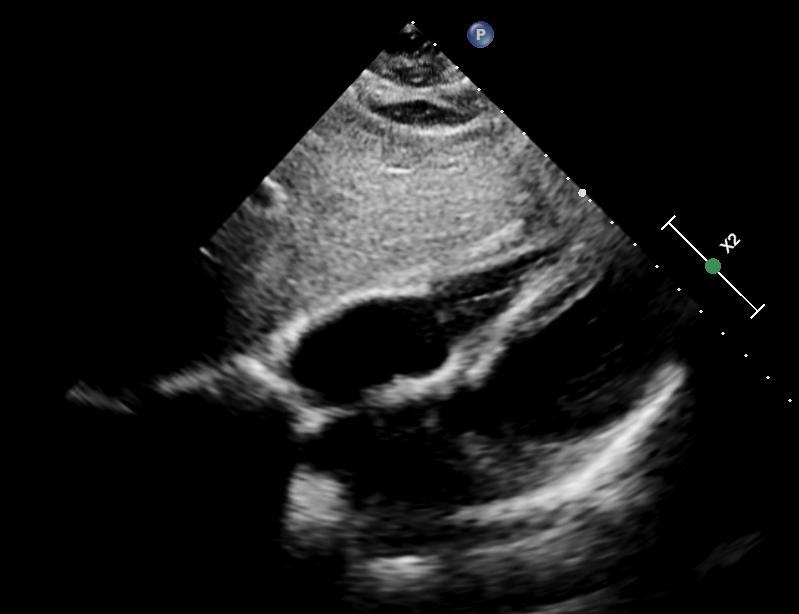}
        \caption{SC4C}
    \end{subfigure}
    
    \caption{Illustration of the nine standard transthoracic echocardiographic views utilized in the EV9V dataset. These views encompass the primary imaging planes for comprehensive cardiac assessment.}
    \label{fig:stand-view}
\end{figure*}

\textbf{Data preprocessing.} To prevent data leakage from correlated videos of the same patient, the dataset was partitioned at the patient level into training, validation, and test sets, comprising 504, 72, and 127 patients, respectively. Detailed statistics are provided in Table~\ref{tab:dataset_distribution}.

All videos were recorded at a uniform frame rate of 30 frames per second, with lengths ranging from 25 to 3,243 frames (median: 116; mean: $177 \pm 190$), reflecting substantial variability in real-world clinical acquisition duration. Frame extraction was performed using FFmpeg (version 4.1), and all extracted frames were resized to a uniform size of $320 \times 240$ pixels. After removing invalid frames, the final dataset contained 910,579 valid image frames spanning nine echocardiographic views (Figure~\ref{fig:stand-view}): (1) Parasternal Long Axis View (PLHLA); (2) Parasternal Short Axis View (PASA); (3) Pulmonary Main Pulmonary Artery Long Axis View (PMPALA); (4) Parasternal Short Axis View at Mitral Valve Level (PMVLSA); (5) Parasternal Short Axis View at Papillary Muscle Level (PPMLSA); (6) Parasternal Short Axis View at Apical Level (PMASA); (7) Apical 4-Chamber View (A4C); (8) Apical 5-Chamber View (A5C); and (9) Subcostal 4-Chamber View (SC4C). These views are standard transthoracic echocardiographic views described in  \citep{mitchell2019guidelines}. Collectively, they encompass key components of routine echocardiographic assessment, including cardiac anatomy, valvular structures, left ventricular function, and intracardiac shunt evaluation.

\subsection{Dataset Characteristics and Statistics}
\begin{figure*}[!tb]
    \centering
    \begin{subfigure}{0.48\textwidth}
        \centering
        \includegraphics[width=\linewidth]{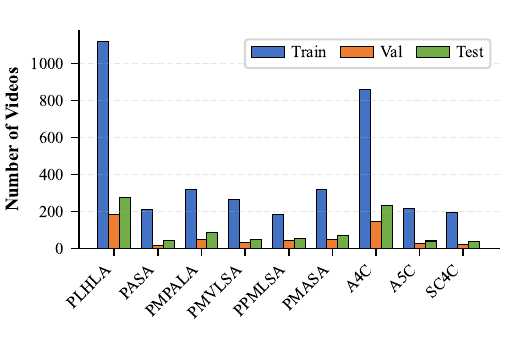}
        \caption{Class distribution across the nine echocardiographic views in the training, validation, and test sets.}
        \label{fig:class_distribution}
    \end{subfigure}
    \hfill
    \begin{subfigure}{0.48\textwidth}
        \centering
        \includegraphics[width=\linewidth]{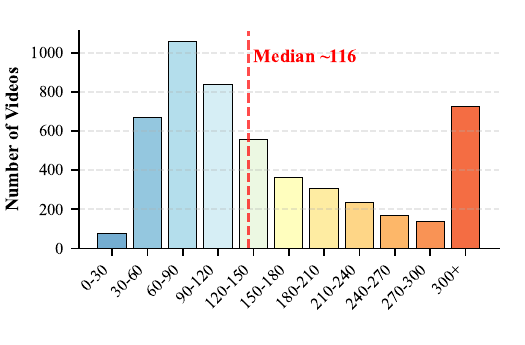}
        \caption{Distribution of video lengths in terms of frame count ranges. The dashed red line indicates the median frame count ($\approx$116).}
        \label{fig:frame_distribution}
    \end{subfigure}
    \caption{Statistical characterization of the EV9V dataset. (a) Class distribution exhibits class imbalance among the nine views, with PLHLA and A4C being the most frequent and PASA among the less represented. (b) Video length distribution shows variability across videos, with frame counts spanning a wide range to reflect differences in clinical scanning duration.}
    \label{fig:dataset_stats}
\end{figure*}

Existing public echocardiographic datasets are limited in both scale and view coverage, hindering the development and clinical translation of view classification algorithms. For example, CAMUS \citep{leclerc2019deep} included only 500 patients with two views (A2C and A4C), while TMED-1 and TMED-2 \citep{huang2022tmed} comprised 260 and 577 patients with four to five views. Moreover, many datasets require formal application or institutional authorization, which limits reproducible research and fair cross-team benchmarking.
\begin{table*}[!tb]
\centering
\caption{Distribution of patients and videos across the training, validation, and test sets, together with the per-view video counts. The first two columns report split statistics; the remaining columns report per-view video counts.}
\label{tab:dataset_distribution}

\begin{tabular}{lccccccccccc}
\toprule
Dataset & Patients & Videos & PLHLA & PASA & PMPALA & PMVLSA & PPMLSA & PMASA & A4C & A5C & SC4C \\
\midrule
Train & 504 & 3,683 & 1,120 & 212 & 319 & 262 & 183 & 317 & 858 & 217 & 195 \\
Val   & 72  & 567   & 185   & 16  & 46  & 32  & 41  & 50  & 148 & 29  & 20  \\
Test  & 127 & 888   & 275   & 41  & 88  & 50  & 56  & 72  & 231 & 40  & 35  \\
\bottomrule
\end{tabular}%

\end{table*}
As shown in Table~\ref{tab:dataset_distribution} and Figure~\ref{fig:dataset_stats}(a), the dataset exhibits class imbalance that reflects real-world clinical acquisition patterns. PLHLA and A4C are the most prevalent views in all splits, while PASA, PMPALA, and SC4C are comparatively underrepresented. This variation naturally arises from clinical practice, where standard views are routinely acquired during every examination, while specialized views require specific patient positioning and are performed only when clinically indicated. Figure~\ref{fig:dataset_stats}(b) further shows the variability in video length across the dataset, with frame counts ranging from 25 to 3,243 (median: 116, mean: $177 \pm 190$), reflecting natural differences in clinical scanning duration. The EV9V dataset exhibits four key characteristics: (1) it is, to our knowledge, the largest publicly accessible dataset, comprising 5,138 videos and 910,579 valid image frames; (2) it provides comprehensive view coverage, encompassing nine clinical standard views including the clinically important yet frequently overlooked PMPALA; (3) it implements rigorous quality control through a three-tier physician annotation system with independent double-blind annotation and senior expert arbitration; and (4) it captures real-world clinical variability in patient anatomy, acoustic windows, and scanning duration for clinically meaningful evaluation.

\section{Proposed Method}
\label{sec:Method}

\subsection{Spatio-Temporal Fusion Model (STFM)}
\label{sec:stfm_overview}

Many standard echocardiographic views exhibit highly similar spatial appearances and can only be reliably distinguished through temporal motion patterns. Existing spatio-temporal modeling approaches have notable limitations: optical flow is sensitive to ultrasound speckle noise and incurs substantial computational overhead \citep{10041980,feichtenhofer2019slowfast,meunier2002echographic}, while 3D convolutional networks introduce large numbers of additional parameters \citep{7410867}.

\begin{figure*}[!tb]
\centering
\includegraphics[width=\textwidth]{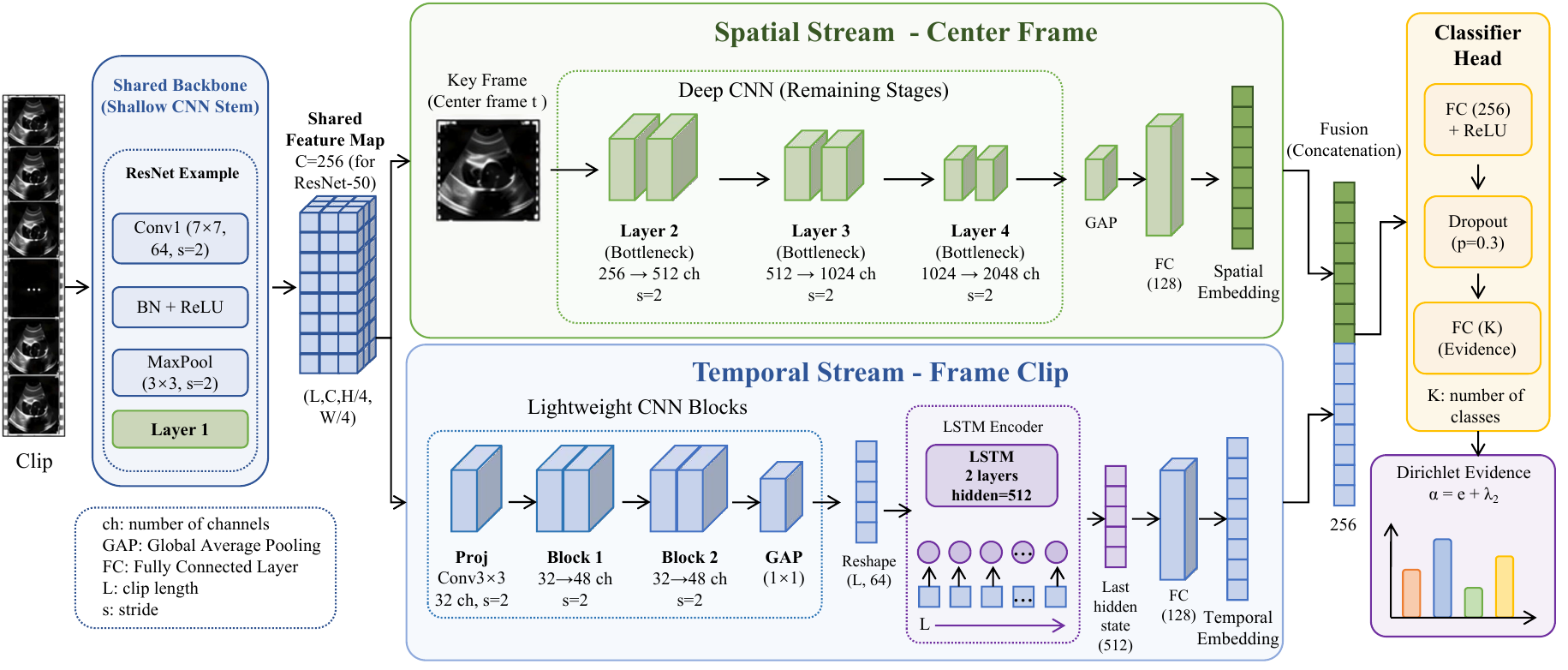}
\caption{Overview of the proposed Spatio-Temporal Fusion Model (STFM). An anchor frame is used for spatial feature extraction, while a short clip centered at the same frame is used for temporal modeling. The two feature streams share a shallow CNN stem and are fused through feature concatenation followed by an MLP. The fused representation is then converted into evidential predictions by the downstream classification head.}
\label{fig:total_framework}
\end{figure*}

To address these issues, we propose the Spatio-Temporal Fusion Model (STFM), a lightweight dual-stream architecture consisting of a spatial branch and a temporal branch (Figure~\ref{fig:total_framework}). Given an echocardiographic video clip, a center frame is fed into the spatial branch to extract anatomical features, while a sparsely sampled temporal clip of length $L$ with a fixed sampling interval $I$, centered at the same frame, is processed by the temporal branch to capture cardiac motion patterns. To reduce computational redundancy, all frames are first passed through a shared shallow CNN stem to extract low-level features. The spatial branch forwards the anchor-frame feature through deeper convolutional layers to generate a spatial embedding. In parallel, the temporal branch processes the features of all sampled frames using a lightweight CNN head followed by a multi-layer LSTM, producing a temporal embedding of the same dimensionality.

Finally, the spatial and temporal embeddings are concatenated and fused through a compact multilayer perceptron to obtain a unified spatio-temporal representation $\mathbf{z}$. This representation jointly encodes anatomical appearance and motion dynamics and serves as the input to the subsequent evidential classification module.

\subsection{Evidential Uncertainty Modeling}
\label{sec:evidential_modeling}

Echocardiographic videos frequently contain frames with varying diagnostic quality and representativeness, including blurred frames, transitional anatomical views, and observations acquired under suboptimal acoustic windows. When multiple clips are aggregated to obtain a video-level prediction, these non-prototypical observations may introduce unreliable evidence and adversely affect the final classification result. Therefore, a reliable mechanism is required to quantify the trustworthiness of individual observations and reduce the influence of ambiguous clips during prediction. To address this issue, we adopt Re-EDL \citep{chen2024redl} as the classification head of STFM. Unlike conventional Softmax classifiers that directly estimate class probabilities, Re-EDL models the amount of evidence supporting each class and represents predictions using a Dirichlet distribution. As a result, representative observations tend to produce stronger evidential support. During inference, predictions from multiple clips are aggregated through evidence fusion, allowing clips with stronger evidential support to contribute more substantially to the final decision.

Following Re-EDL, the network outputs a non-negative evidence vector $\mathbf{e} = [e_1,\ldots,e_K]$ via a softplus activation, which parameterizes a Dirichlet distribution with concentration parameters $\alpha_k = e_k + \lambda_2$. Predictive uncertainty is then quantified as
\begin{equation}
\label{eq:uncertainty_u}
u = \frac{\lambda_2 K}{\sum_{j=1}^{K}(e_j + \lambda_2)},
\end{equation}
and 
\begin{equation}
\hat{p}_k = \frac{e_k + \lambda_2}{e_k + \lambda_1 (\mathcal{E} - e_k) + \lambda_2 K},
\label{eq:reedl_prob}
\end{equation}
where $\mathcal{E} = \sum_{j=1}^{K} e_j$ and $K$ is the number of classes. Following \citep{chen2024redl}, the prior-strength parameters are set to $\lambda_1 = 1.0$ and $\lambda_2 = 0.8$. Larger uncertainty values correspond to weaker evidential support. The network is optimized by minimizing the mean squared error between the projected probabilities $\hat{\mathbf{p}}$ and one-hot ground-truth labels.

During inference, $T$ key frames are uniformly sampled across the entire video to provide comprehensive temporal coverage. For each sampled frame $t$, STFM processes the corresponding local clip and produces an evidence vector $\mathbf{e}^{(t)}$ together with the associated Dirichlet parameters $\boldsymbol{\alpha}^{(t)}$. The video-level Dirichlet distribution is obtained by pooling the frame-level evidence and assigning the uniform prior once to the pooled quantity:
\begin{equation}
\boldsymbol{\alpha}^{(video)}
=
\sum_{t=1}^{T}
 \mathbf{e}^{(t)}
+\;\lambda_2\mathbf{1},
\label{eq:alpha_fusion}
\end{equation}
where $\boldsymbol{\alpha}^{(video)}$ represents the pooled evidence and prior from all sampled observations, and $\mathbf{1}$ is the vector of ones over the $K$ classes. The probability assigned to class $k$ for the video is taken as the mean of this pooled Dirichlet distribution, $\hat{p}_k^{(video)}=\alpha_k^{(video)}/\sum_{j=1}^{K}\alpha_j^{(video)}$, and it is this quantity whose calibration is measured in Section~\ref{sec:calibration}.

\subsection{Uncertainty-Guided Segment Selection}
\label{sec:segment_selection}

However, the effectiveness of evidence-based fusion depends on the model's ability to assign distinct evidential strengths to representative and non-representative observations. Excessive optimization on ambiguous or low-quality samples may reduce this distinction and weaken the benefits of uncertainty-aware fusion. To preserve the discriminative capability of Re-EDL, we further leverage its uncertainty estimates to guide training sample selection. Specifically, video segments that consistently exhibit lower uncertainty are regarded as more representative and are therefore sampled more frequently during training.

Each video $V_i$ containing $n_i$ frames is partitioned into $m_i=\lceil n_i/L_s\rceil$ non-overlapping segments of length $L_s$. The last segment may contain fewer frames if $n_i$ is not divisible by $L_s$. Rather than maintaining certainty scores for individual frames, certainty is estimated at the segment level because adjacent echocardiographic frames are highly correlated and typically share similar anatomical content. Segment-level modeling therefore provides a more stable estimate of temporal reliability while substantially reducing the amount of certainty information that must be maintained during training.

For each segment $s$, a certainty score $c_{i,s}$ is maintained in a global segment bank $\mathcal{C}$, where larger values indicate a higher likelihood that the segment contains representative in-distribution observations. During training, segment selection and score updates are both performed based on the uncertainty estimates obtained from sampled clips. An anchor frame is then randomly sampled from the selected segment, and a temporal clip centered at this frame is constructed for model training. Based on the uncertainty mass $u$ obtained from Eq.~(\ref{eq:uncertainty_u}), the instantaneous certainty from the current forward pass is defined as
\begin{equation}
c^{(new)} = 1 - u, \qquad u \in [0,1].
\end{equation}

After each training iteration, the certainty scores of the sampled segments are updated using an exponential moving average (EMA):
\begin{equation}
c_{i,s}^{(t+1)}=\eta c^{(new)}+(1-\eta)c_{i,s}^{(t)},
\end{equation}

where $\eta=0.3$ denotes the EMA momentum and $c^{(new)}$ is the certainty computed from the current model output. To distinguish unexplored segments from previously visited ones, all entries in the segment bank are initialized to $-0.1$ as a sentinel value. Upon the first visit, the certainty score is assigned directly rather than updated via EMA.

To balance exploitation of representative segments and exploration of the entire video, an $\varepsilon$-greedy strategy is adopted. For each training iteration on video $V_i$, a target segment $s^*$ is selected according to
\begin{equation}
s^*=
\begin{cases}
\text{random } s \in \{1,\ldots,m_i\},
&
\text{with probability } \varepsilon,
\\
\arg\max_s \mathcal{C}[i][s],
&
\text{with probability } 1-\varepsilon,
\end{cases}
\end{equation}
where the exploration probability is set to $\varepsilon=$ $0.8$. With probability $1-\varepsilon$, the training process preferentially selects the segment with the highest historical certainty, encouraging learning from representative observations. With probability $\varepsilon$, a random segment is selected to ensure sufficient exploration and prevent the optimization process from collapsing onto a small subset of segments.

Once the target segment $s^*$ is determined, a single frame is uniformly sampled from the segment and used as the anchor frame for subsequent spatio-temporal feature extraction. The complete segment-selection and certainty-update procedure is summarized in Algorithm~\ref{alg:selective_training}.
\begin{algorithm}[htbp]
\caption{Uncertainty-Guided Segment Selection}
\label{alg:selective_training}
\begin{algorithmic}[1]
\REQUIRE
Video $V_i$ partitioned into $m_i$ segments,
segment bank $\mathcal{C}$,
exploration probability $\varepsilon$,
EMA momentum $\eta$

\FOR{video $V_i$}
    \IF{rand() $< \varepsilon$}
        \STATE $s^* \sim \mathrm{Uniform}(\{1,\ldots,m_i\})$
    \ELSE
        \STATE $s^* \leftarrow \arg\max_s \mathcal{C}[i][s]$
    \ENDIF

    \STATE Sample anchor frame $f^*$ from segment $s^*$

    \STATE Generate temporal clip centered at $f^*$

    \STATE Compute uncertainty $u$ using Eq.~(\ref{eq:uncertainty_u})

    \STATE $c^{(new)} \leftarrow 1 - u$

    \IF{$\mathcal{C}[i][s^*] = -0.1$}
        \STATE $\mathcal{C}[i][s^*] \leftarrow c^{(new)}$
    \ELSE
        \STATE $\mathcal{C}[i][s^*] \leftarrow \eta c^{(new)} + (1-\eta)\mathcal{C}[i][s^*]$
    \ENDIF
\ENDFOR
\end{algorithmic}
\end{algorithm}
This inference strategy complements the uncertainty-guided segment selection employed during training. By preferentially sampling representative observations, the proposed training procedure helps preserve the evidential model's ability to distinguish well-supported in-distribution samples from non-prototypical observations. Consequently, observations that produce stronger evidence during inference naturally contribute more to the fused prediction, establishing consistency between the training and inference stages.

\section{Experiments}
\label{sec:baseline_evaluation}

This section evaluates a diverse set of methods on EV9V, including conventional CNNs (ResNet, EfficientNet, ConvNeXt), transformers (ViT, Swin Transformer), and video architectures (I3D, TSN, SlowOnly, UniFormerV2), to establish a comprehensive benchmark for echocardiographic view classification.

\paragraph{Experimental environment.} All experiments involving the proposed STFM, including its main configuration and all ablation studies, were conducted on a workstation equipped with two NVIDIA GeForce RTX 5090 GPUs (32 GB memory each) and two Intel Xeon Gold 6459C CPUs (32 cores and 64 threads each), running Ubuntu 22.04.5 LTS. The software environment comprises Python 3.12.4, PyTorch 2.8.0 built against CUDA 12.8 with cuDNN 9.10, and torchvision 0.23.0. The frame-level 2D baselines and the compared video-level architectures were trained on a separate machine equipped with two NVIDIA GeForce RTX 4090 GPUs (24 GB memory each) and an AMD EPYC 7T83 CPU (16 cores). The software environment for these experiments comprises Python 3.11 and PyTorch 2.9.1 built against CUDA 12.1.1.

\subsection{2D Frame-level Baselines}

\paragraph{Training Protocol}
We adopt a frame-level training with video-level aggregation paradigm. All videos are decomposed into individual frames, yielding 910,579 frames in total. At each training step, one random frame is sampled per video and the model is trained on single-frame inputs. During testing, 30 equally spaced frames are sampled from each video and processed independently; the final video-level prediction is obtained via majority voting. This protocol follows prior frame-based echocardiographic view classification studies and provides a simple video-level aggregation baseline. From these predictions we compute Top-1 accuracy, macro-averaged recall, and macro-averaged F1 score (Macro F1) at the video level.

\paragraph{Implementation Details}
Data augmentation includes random rotation ($\pm$15 degrees), random resized cropping (scale 0.8 to 1.0), random horizontal flipping (probability 0.5), and random Gaussian blurring (kernel size 3, sigma range 0.1 to 2.0, probability 0.5). Input dimensions follow two sizes:\\ 224$\times$224 for ResNet, ConvNeXt, ViT, and Swin Transformer; 256$\times$256 for EfficientNet-B7, EfficientNet-V2-L, and Swin-V2-B. Pixel normalization uses ImageNet statistics (mean $=[0.485, 0.456, 0.406]$, std $=[0.229, 0.224, 0.225]$) to remain compatible with pretrained weights.

Training runs for 100 epochs using the Adam optimizer with a learning rate of $10^{-4}$ and weight decay of 0.01. The learning rate decays by a factor of 0.1 every 10 epochs via StepLR. All models are initialized with ImageNet pretrained weights and fine-tuned end-to-end.

\paragraph{Results and Analysis}
Table~\ref{tab:2d_baseline} summarizes the performance of representative 2D architectures on EV9V. RegNet achieves the best test accuracy (93.32\%) and macro F1 (89.02\%), closely followed by ResNet-18 (92.98\% / 88.43\%) and EfficientNet-B7 (92.91\% / 88.53\%). Overall, modern CNN architectures consistently achieve strong performance, with over half of the models exceeding 92\% test accuracy.

\begin{table*}[!tb]
\centering
\caption{2D frame-level baseline results on EV9V. All metrics are reported in percentage (\%).}
\label{tab:2d_baseline}
\setlength{\tabcolsep}{10pt}
\begin{tabular}{lcc@{\hskip 6pt}ccc@{\hskip 4pt}ccc}
\toprule
\multirow{2}{*}{Model} & \multirow{2}{*}{Input} & \multicolumn{3}{c}{Validation} & \multicolumn{3}{c}{Test} \\
\cmidrule{3-5} \cmidrule{6-8}
 & & Acc. & Recall & F1 & Acc. & Recall & F1 \\
\midrule
ResNet-18 \citep{he2016deep}  & 224 & $93.42 {\scriptscriptstyle \pm 0.10}$ & $88.00 {\scriptscriptstyle \pm 0.18}$ & $88.23 {\scriptscriptstyle \pm 0.09}$ & $92.98 {\scriptscriptstyle \pm 0.43}$ & $87.84 {\scriptscriptstyle \pm 1.04}$ & $88.43 {\scriptscriptstyle \pm 1.08}$ \\
ResNet-50 \citep{he2016deep}  & 224 & $92.77 {\scriptscriptstyle \pm 0.47}$ & $85.91 {\scriptscriptstyle \pm 0.87}$ & $87.28 {\scriptscriptstyle \pm 1.00}$ & $92.53 {\scriptscriptstyle \pm 0.68}$ & $87.06 {\scriptscriptstyle \pm 1.48}$ & $87.91 {\scriptscriptstyle \pm 1.44}$ \\
\midrule
DenseNet-121 \citep{huang2017densely} & 224 & $\mathbf{93.89 {\scriptscriptstyle \pm 0.80}}$ & $\mathbf{88.59 {\scriptscriptstyle \pm 1.72}}$ & $\mathbf{89.25 {\scriptscriptstyle \pm 1.37}}$ & $92.61 {\scriptscriptstyle \pm 0.40}$ & $87.84 {\scriptscriptstyle \pm 1.58}$ & $88.27 {\scriptscriptstyle \pm 0.99}$ \\
DenseNet-169 \citep{huang2017densely} & 224 & $92.83 {\scriptscriptstyle \pm 0.27}$ & $86.02 {\scriptscriptstyle \pm 0.48}$ & $87.14 {\scriptscriptstyle \pm 0.64}$ & $92.34 {\scriptscriptstyle \pm 0.88}$ & $86.43 {\scriptscriptstyle \pm 1.57}$ & $87.37 {\scriptscriptstyle \pm 1.51}$ \\
\midrule
EfficientNet-B7 \citep{tan2019efficientnet}  & 256 & $93.24 {\scriptscriptstyle \pm 0.57}$ & $88.40 {\scriptscriptstyle \pm 1.26}$ & $87.98 {\scriptscriptstyle \pm 1.08}$ & $92.91 {\scriptscriptstyle \pm 0.30}$ & $\mathbf{88.45 {\scriptscriptstyle \pm 0.33}}$ & $88.53 {\scriptscriptstyle \pm 0.52}$ \\
EfficientNetV2-L \citep{tan2021efficientnetv2} & 256 & $93.42 {\scriptscriptstyle \pm 0.37}$ & $87.98 {\scriptscriptstyle \pm 0.76}$ & $88.08 {\scriptscriptstyle \pm 0.57}$ & $92.68 {\scriptscriptstyle \pm 0.88}$ & $87.90 {\scriptscriptstyle \pm 1.55}$ & $88.19 {\scriptscriptstyle \pm 1.62}$ \\
\midrule
ConvNeXt-Tiny \citep{liu2022convnet}  & 224 & $92.89 {\scriptscriptstyle \pm 0.20}$ & $86.36 {\scriptscriptstyle \pm 0.92}$ & $87.31 {\scriptscriptstyle \pm 0.90}$ & $92.08 {\scriptscriptstyle \pm 0.69}$ & $85.85 {\scriptscriptstyle \pm 1.10}$ & $86.81 {\scriptscriptstyle \pm 1.34}$ \\
ConvNeXt-Small \citep{liu2022convnet} & 224 & $93.53 {\scriptscriptstyle \pm 0.20}$ & $87.20 {\scriptscriptstyle \pm 0.50}$ & $88.10 {\scriptscriptstyle \pm 0.32}$ & $92.49 {\scriptscriptstyle \pm 0.62}$ & $87.09 {\scriptscriptstyle \pm 1.31}$ & $87.76 {\scriptscriptstyle \pm 1.02}$ \\
ConvNeXt-Base \citep{liu2022convnet}  & 224 & $92.59 {\scriptscriptstyle \pm 0.18}$ & $86.02 {\scriptscriptstyle \pm 0.52}$ & $86.86 {\scriptscriptstyle \pm 0.34}$ & $91.97 {\scriptscriptstyle \pm 0.40}$ & $86.28 {\scriptscriptstyle \pm 1.03}$ & $86.95 {\scriptscriptstyle \pm 0.77}$ \\
\midrule
MobileNetV3-Small \citep{Koonce2021} & 224 & $88.65 {\scriptscriptstyle \pm 0.10}$ & $78.84 {\scriptscriptstyle \pm 0.76}$ & $79.24 {\scriptscriptstyle \pm 1.62}$ & $88.85 {\scriptscriptstyle \pm 0.56}$ & $79.95 {\scriptscriptstyle \pm 1.88}$ & $80.56 {\scriptscriptstyle \pm 1.97}$ \\
MobileNetV3-Large \citep{Koonce2021} & 224 & $90.89 {\scriptscriptstyle \pm 0.10}$ & $82.30 {\scriptscriptstyle \pm 0.52}$ & $83.56 {\scriptscriptstyle \pm 0.60}$ & $90.20 {\scriptscriptstyle \pm 0.63}$ & $82.46 {\scriptscriptstyle \pm 1.26}$ & $83.59 {\scriptscriptstyle \pm 1.35}$ \\
\midrule
RegNet \citep{radosavovic2020designing} & 224 & $\mathbf{93.89 {\scriptscriptstyle \pm 0.37}}$ & $88.08 {\scriptscriptstyle \pm 0.68}$ & $88.98 {\scriptscriptstyle \pm 0.54}$ & $\mathbf{93.32 {\scriptscriptstyle \pm 0.47}}$ & $88.18 {\scriptscriptstyle \pm 0.33}$ & $\mathbf{89.02 {\scriptscriptstyle \pm 0.61}}$ \\
\midrule
ViT-L-16 \citep{dosovitskiyimage} & 224 & $92.00 {\scriptscriptstyle \pm 0.57}$ & $85.47 {\scriptscriptstyle \pm 0.96}$ & $86.37 {\scriptscriptstyle \pm 1.14}$ & $90.84 {\scriptscriptstyle \pm 0.13}$ & $84.16 {\scriptscriptstyle \pm 1.03}$ & $84.58 {\scriptscriptstyle \pm 0.80}$ \\
MaxViT-T \citep{tu2022maxvit} & 224 & $89.07 {\scriptscriptstyle \pm 0.00}$ & $80.04 {\scriptscriptstyle \pm 0.29}$ & $81.18 {\scriptscriptstyle \pm 0.35}$ & $88.74 {\scriptscriptstyle \pm 1.30}$ & $80.12 {\scriptscriptstyle \pm 2.25}$ & $81.21 {\scriptscriptstyle \pm 2.17}$ \\
Swin-B \citep{liu2021swin}   & 224 & $92.36 {\scriptscriptstyle \pm 0.67}$ & $85.00 {\scriptscriptstyle \pm 1.49}$ & $86.24 {\scriptscriptstyle \pm 0.98}$ & $91.22 {\scriptscriptstyle \pm 0.23}$ & $84.61 {\scriptscriptstyle \pm 0.80}$ & $85.64 {\scriptscriptstyle \pm 0.48}$ \\
SwinV2-B \citep{liu2022swin} & 256 & $92.48 {\scriptscriptstyle \pm 0.10}$ & $85.94 {\scriptscriptstyle \pm 0.92}$ & $86.62 {\scriptscriptstyle \pm 0.08}$ & $92.08 {\scriptscriptstyle \pm 0.17}$ & $86.73 {\scriptscriptstyle \pm 0.56}$ & $87.13 {\scriptscriptstyle \pm 0.31}$ \\
\bottomrule
\end{tabular}%
\end{table*}

A broader comparison reveals that CNN-based models consistently outperform Transformer-based architectures (ViT, MaxViT, Swin) under the current frame-level setting. One possible explanation is that echocardiographic view classification relies heavily on fine-grained local anatomical structures, which can be effectively captured by convolutional feature extractors. In addition, the frame-level training protocol provides relatively limited supervision per video, potentially limiting the advantages of Transformer architectures, which often benefit from larger-scale training data and stronger supervision.

For CNN-based architectures, macro F1 scores are 4.3--8.3\% lower than accuracy, indicating that class imbalance remains a challenge and disproportionately affects minority-class recall.

The generalization gap between validation and test accuracy is generally small for most architectures, with the largest gaps occurring for DenseNet-121 (1.28\%), ViT-L-16 (1.16\%) and Swin-B (1.14\%).

Nevertheless, the performance differences between CNN and Transformer models remain relatively modest compared with the overall variation observed across different video-level approaches in later sections, suggesting that further performance improvements may depend not only on stronger spatial backbones but also on the effective incorporation of temporal information.

\subsection{Video-level Benchmark}

\paragraph{Evaluated Architectures}
To establish comprehensive video-level baselines on EV9V, we evaluate representative video classification architectures spanning four major paradigms: 2D CNNs with temporal aggregation, RNN-based models, 3D convolutional networks, and video transformers. Specifically, the evaluated models include TSN, TSM, TANet, and TIN for lightweight temporal modeling based on 2D backbones; ResNet-18 with LSTM, GRU and BiLSTM for recurrent temporal modeling; C3D, I3D, R(2+1)D, C2D, SlowOnly, and TPN for 3D convolutional modeling; and TimeSformer, Swin Transformer, and UniFormerV2 for transformer-based video understanding. All implementations are based on the MMAction2 \citep{2020mmaction2} framework with architecture-specific adaptations for echocardiographic videos.

\paragraph{Implementation Details}
Model configurations generally follow the original implementations. For most architectures, the clip length, sampling strategy, optimizer, and training schedule remain unchanged. Notable exceptions include I3D and SlowFast-related models, whose temporal spans are reduced from 64 to 16 frames to better match the duration characteristics of echocardiographic videos. Learning rates and augmentation strategies are adjusted when required by the original implementations, such as ColorJitter for TPN. Each model is trained and evaluated using three random seeds, and we report the mean and standard deviation of the results. Performance is assessed using video-level Top-1 accuracy and macro F1 score, the latter computed from the confusion matrix of the nine views. FLOPs are counted for a single forward pass of each model, covering one complete video-level inference and excluding test-time augmentation; in the backbone ablation, they are reported for a single 5-frame clip processed by the backbone. For the proposed STFM, the default configuration and the training protocol used in all STFM experiments are described in Section~\ref{sec:ablation}.

\paragraph{Results and Analysis}
Table~\ref{tab:video_results} summarizes the benchmark results. Compared with the frame-level baselines in Table~\ref{tab:2d_baseline}, the video architectures are on average clearly stronger (mean test accuracy 93.5\% against 91.7\%), although six of them---TimeSformer, R(2+1)D, TIN, BiLSTM, C2D and C3D---remain below the best frame-level model, RegNet (93.32\%), so temporal information helps but a video backbone alone is not sufficient. Among all evaluated architectures, the proposed STFM (instantiated with a ResNet-18 backbone) achieves a test accuracy (93.99\%) comparable to that of the best-performing video transformer UniFormerV2 (94.41\%). The gap is smaller than the seed-to-seed standard deviation reported for a single configuration. Notably, STFM requires substantially fewer parameters (14.75M vs.\ 114.25M) and computational cost (43.13G vs.\ 570.23G FLOPs, both covering one full video-level inference), highlighting a favorable accuracy--efficiency trade-off. Performance differences across model families reveal distinct characteristics. Lightweight temporal modeling methods based on 2D CNN backbones consistently achieve strong results while maintaining low computational complexity. Transformer-based architectures remain competitive, with UniFormerV2 substantially outperforming TimeSformer, suggesting that convolutional inductive biases remain beneficial at the current dataset scale. Meanwhile, 3D convolutional networks generally outperform frame-level models but do not consistently surpass the strongest temporal modeling approaches. These results suggest that effective temporal modeling is more critical than increasing model complexity alone, as lightweight architectures can match or even outperform substantially larger video models on EV9V. Because the models at the top of Table~\ref{tab:video_results} differ by less than a percentage point, we also examined how large such a difference has to be before it can be read as a gap between models. The 95\% Wilson interval~\citep{wilson1927probable} of the test accuracy covers 92.23--95.38\% for STFM, 92.49--95.57\% for TPN and 92.53--95.60\% for UniFormerV2, so the differences among them are not large enough to be clearly resolved on the 888 test videos. A paired exact McNemar test~\citep{mcnemar1947note} on the same videos, comparing each of the three seeds of STFM with the corresponding run of the six strongest baselines, which gives 18 tests in all, finds no comparison in which a baseline is significantly ahead after Holm correction across the 18 tests; the smallest $p$-value among the three tests against UniFormerV2, the most accurate baseline in Table~\ref{tab:video_results}, is 0.16.

\begin{table*}[!tb]
\centering

\caption{Video model evaluation results on EV9V. Models are grouped by architecture type. Macro F1 is computed from confusion matrices. Each model is evaluated with three random seeds. All metrics are reported in percentage (\%).}
\label{tab:video_results}
\setlength{\tabcolsep}{5pt}
\begin{tabular}{lcc@{\hskip 4pt}c@{\hskip 6pt}cccc}
\toprule
\multirow{2}{*}{Model} & \multirow{2}{*}{Params} & \multirow{2}{*}{Input} & \multirow{2}{*}{FLOPs} & \multicolumn{2}{c}{Val} & \multicolumn{2}{c}{Test} \\
\cmidrule{5-8}
 & & & & Acc. & F1 & Acc. & F1 \\
\midrule
\multicolumn{8}{l}{\textbf{2D CNN + Temporal Aggregation}} \\
TSN (R50) \citep{wang2016temporal}          & 23.53M   & $8\times3\times224^2$ & 32.87G  & $95.53 {\scriptscriptstyle \pm 0.54}$ & $92.31 {\scriptscriptstyle \pm 1.35}$ & $93.99 {\scriptscriptstyle \pm 0.07}$ & $90.31 {\scriptscriptstyle \pm 0.20}$ \\
TSM (R50) \citep{lin2019tsm}          & 23.53M   & $8\times3\times224^2$ & 32.87G  & $94.47 {\scriptscriptstyle \pm 0.27}$ & $89.97 {\scriptscriptstyle \pm 0.61}$ & $93.66 {\scriptscriptstyle \pm 0.90}$ & $89.95 {\scriptscriptstyle \pm 1.43}$ \\
TANet (R50) \citep{liu2021tam}        & 24.79M   & $8\times3\times224^2$ & 32.95G  & $94.30 {\scriptscriptstyle \pm 0.37}$ & $89.65 {\scriptscriptstyle \pm 0.62}$ & $94.07 {\scriptscriptstyle \pm 0.72}$ & $90.45 {\scriptscriptstyle \pm 1.14}$ \\
TIN (R50) \citep{shao2020temporal}          & 23.56M   & $8\times3\times224^2$ & 32.87G  & $94.00 {\scriptscriptstyle \pm 0.61}$ & $88.91 {\scriptscriptstyle \pm 0.97}$ & $93.21 {\scriptscriptstyle \pm 0.77}$ & $89.16 {\scriptscriptstyle \pm 1.49}$ \\
\midrule
\multicolumn{8}{l}{\textbf{RNN-based}} \\
ResNet18 + LSTM   & 15.38M & $16\times3\times224^2$ & 58.49G & $94.06 {\scriptscriptstyle \pm 0.62}$ & $89.21 {\scriptscriptstyle \pm 1.25}$ & $93.92 {\scriptscriptstyle \pm 0.41}$ & $90.54 {\scriptscriptstyle \pm 0.47}$ \\
ResNet18 + GRU    & 14.33M & $16\times3\times224^2$ & 58.45G & $94.71 {\scriptscriptstyle \pm 0.61}$ & $90.54 {\scriptscriptstyle \pm 1.44}$ & $93.54 {\scriptscriptstyle \pm 0.28}$ & $90.10 {\scriptscriptstyle \pm 0.31}$ \\
ResNet18 + BiLSTM & 21.69M & $16\times3\times224^2$ & 58.69G & $93.94 {\scriptscriptstyle \pm 0.27}$ & $89.15 {\scriptscriptstyle \pm 1.05}$ & $93.17 {\scriptscriptstyle \pm 0.65}$ & $89.28 {\scriptscriptstyle \pm 0.95}$ \\
\midrule
\multicolumn{8}{l}{\textbf{Video Transformer}} \\
TimeSformer \citep{bertasius2021space}        & 121.27M  & $8\times3\times224^2$ & 196.05G & $92.83 {\scriptscriptstyle \pm 0.27}$ & $87.10 {\scriptscriptstyle \pm 0.49}$ & $91.52 {\scriptscriptstyle \pm 0.43}$ & $86.35 {\scriptscriptstyle \pm 0.86}$ \\
Swin-T (P244) \citep{liu2021video}      & 27.86M   & $128\times3\times224^2$ & 352.25G & $94.71 {\scriptscriptstyle \pm 1.22}$ & $90.69 {\scriptscriptstyle \pm 1.98}$ & $93.54 {\scriptscriptstyle \pm 0.85}$ & $89.95 {\scriptscriptstyle \pm 1.43}$ \\
UniFormerV2 (B/16) \citep{li2022uniformerv2} & 114.25M  & $32\times3\times224^2$ & 570.23G & $\mathbf{95.83 {\scriptscriptstyle \pm 0.10}}$ & $\mathbf{92.56 {\scriptscriptstyle \pm 0.38}}$ & $\mathbf{94.41 {\scriptscriptstyle \pm 0.43}}$ & $\mathbf{91.14 {\scriptscriptstyle \pm 0.51}}$ \\
\midrule
\multicolumn{8}{l}{\textbf{3D CNN}} \\
C3D \citep{7410867}                & 78.03M   & $160\times3\times112^2$ & 385.47G & $94.12 {\scriptscriptstyle \pm 0.37}$ & $89.40 {\scriptscriptstyle \pm 0.55}$ & $92.83 {\scriptscriptstyle \pm 0.46}$ & $88.78 {\scriptscriptstyle \pm 0.50}$ \\
I3D (R50) \citep{carreira2017quo}          & 27.24M   & $160\times3\times256^2$ & 217.28G & $94.36 {\scriptscriptstyle \pm 0.31}$ & $89.80 {\scriptscriptstyle \pm 0.48}$ & $93.32 {\scriptscriptstyle \pm 0.36}$ & $89.26 {\scriptscriptstyle \pm 0.62}$ \\
R(2+1)D (R34) \citep{Tran2017ACL}      & 63.76M   & $80\times3\times256^2$ & 531.41G & $94.65 {\scriptscriptstyle \pm 0.71}$ & $89.90 {\scriptscriptstyle \pm 0.37}$ & $93.28 {\scriptscriptstyle \pm 0.34}$ & $89.22 {\scriptscriptstyle \pm 0.45}$ \\
C2D (R50) \citep{wang2018non}          & 23.53M   & $80\times3\times224^2$ & 194.17G & $94.53 {\scriptscriptstyle \pm 0.71}$ & $90.08 {\scriptscriptstyle \pm 1.43}$ & $92.91 {\scriptscriptstyle \pm 0.52}$ & $88.72 {\scriptscriptstyle \pm 0.77}$ \\
SlowOnly (R50) \citep{feichtenhofer2019slowfast}     & 31.65M   & $80\times3\times256^2$ & 547.50G & $95.47 {\scriptscriptstyle \pm 0.44}$ & $92.19 {\scriptscriptstyle \pm 0.68}$ & $93.92 {\scriptscriptstyle \pm 0.23}$ & $90.22 {\scriptscriptstyle \pm 0.21}$ \\
TPN (R50) \citep{yang2020tpn}    & 91.50M   & $80\times3\times256^2$ & 547.50G & $94.42 {\scriptscriptstyle \pm 0.57}$ & $90.21 {\scriptscriptstyle \pm 1.00}$ & $94.22 {\scriptscriptstyle \pm 0.33}$ & $90.70 {\scriptscriptstyle \pm 0.64}$ \\
\midrule
Ours & 14.75M & $50\times3\times224^2$ & 43.13G & $95.00 {\scriptscriptstyle \pm 0.10}$ & $90.84 {\scriptscriptstyle \pm 0.28}$ & $93.99 {\scriptscriptstyle \pm 1.02}$ & $90.24 {\scriptscriptstyle \pm 1.74}$ \\
\bottomrule
\end{tabular}%
\end{table*}

\section{Ablation Study on STFM Components}
\label{sec:ablation}

We conduct ablation experiments to analyze the contribution of each STFM component. Each configuration is evaluated with three random seeds (100, 200, 300) and we report mean $\pm$ std of accuracy and macro F1 on both validation and test sets. Unless otherwise specified, all STFM models are trained with the Adam optimizer for up to 100 epochs using a batch size of 64, a learning rate of $2\times10^{-5}$, and a weight decay of 0.05. A 5-epoch warmup schedule is applied to the exploration probability $\varepsilon$ at the beginning of training, and no cosine annealing is used. Data augmentation applies random resized cropping and random rotation, shared within each clip but independent across clips, together with per-frame brightness and contrast jittering; class-balanced oversampling is used during training. Training stops early with a patience of 10 epochs on validation macro F1. At inference, $T=10$ key frames are uniformly sampled from each video. The default configuration uses a ResNet-18 backbone with LSTM hidden size 512 and 2 layers, clip length $L=5$ and sampling interval $I=5$. Every choice above is made on validation metrics, which also govern checkpoint selection during training; test metrics are reported post hoc, and differences comparable to the seed-to-seed deviation are not treated as meaningful. The ablation tables in this section come from three independent sweeps run on different days; re-running the same nominal configuration shifts validation accuracy by up to about 0.3 percentage points, so its absolute values may differ slightly from those in Tables~\ref{tab:video_results}, \ref{tab:ablation_epsilon} and~\ref{tab:data_scaling}. Every comparison below is made within a single sweep. All ablation results use the configuration selected on the validation split ($\varepsilon=0.8$, $\lambda_2=0.8$, hidden size 512 with two layers); the tables in this section also contain settings that score higher on the test split alone (94.48\% at $\varepsilon=0.4$), which we do not report as the result of the model, because selecting the configuration on the test split would invalidate the comparison with the baselines.

\begin{figure*}[!tb]
    \centering
    \begin{subfigure}{0.637\textwidth}
        \centering
        \setlength{\tabcolsep}{3pt}\begin{tabular}{lc@{\hskip 4pt}ccc@{\hskip 4pt}ccc@{\hskip 4pt}ccc}
        \toprule
         &  & \multicolumn{3}{c}{Precision} & \multicolumn{3}{c}{Recall} & \multicolumn{3}{c}{F1} \\
        \cmidrule(l){3-5} \cmidrule(l){6-8} \cmidrule(l){9-11}
        View & Sup. & Reg. & Unif. & STFM & Reg. & Unif. & STFM & Reg. & Unif. & STFM \\
        \midrule
        PLHLA & 275 & 100.0 & 100.0 & 100.0 & 100.0 & 100.0 & 100.0 & 100.0 & 100.0 & 100.0 \\
        \addlinespace[3pt]
        PMASA & 72 & 84.4 & 83.1 & 88.2 & 86.1 & 86.1 & 84.3 & 85.2 & 84.4 & 85.9 \\
        \addlinespace[3pt]
        PMVLSA & 50 & 86.2 & 96.1 & 95.2 & 95.3 & 96.0 & 91.3 & 90.5 & 96.0 & 93.2 \\
        \addlinespace[3pt]
        PASA & 41 & 85.4 & 87.5 & 83.9 & 80.5 & 79.7 & 80.5 & 82.8 & 83.4 & 82.2 \\
        \addlinespace[3pt]
        A4C & 231 & 95.2 & 97.4 & 98.0 & 99.3 & 98.8 & 97.3 & 97.2 & 98.1 & 97.6 \\
        \addlinespace[3pt]
        A5C & 40 & 93.6 & 90.3 & 83.8 & 71.7 & 85.0 & 86.7 & 80.9 & 87.5 & 85.1 \\
        \addlinespace[3pt]
        PMPALA & 88 & 87.3 & 88.2 & 86.8 & 85.6 & 84.1 & 89.4 & 86.4 & 85.9 & 87.9 \\
        \addlinespace[3pt]
        PPMLSA & 56 & 82.4 & 83.0 & 81.0 & 78.0 & 89.9 & 86.3 & 80.1 & 86.3 & 83.6 \\
        \addlinespace[3pt]
        SC4C & 35 & 99.0 & 100.0 & 96.2 & 97.1 & 97.1 & 97.1 & 98.1 & 98.6 & 96.7 \\
        \addlinespace[3pt]
        Macro & 888 & 90.4 & 91.8 & 90.4 & 88.2 & 90.7 & 90.3 & 89.0 & 91.1 & 90.2 \\
        \bottomrule
        \end{tabular}
    \caption{Per-class precision, recall and F1 score (\%) on the EV9V test set for RegNet (Reg.), UniFormerV2 (Unif.) and the proposed STFM.}
        \label{subfig:perclass_table}
    \end{subfigure}
    \hfill
    \begin{subfigure}{0.339\textwidth}
        \centering
        \includegraphics[width=\linewidth]{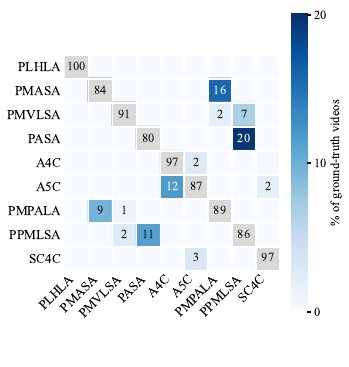}
    \caption{Confusion matrix of the proposed STFM on the same test videos.}
        \label{subfig:perclass_cm}
    \end{subfigure}
    \caption{Per-class behaviour on the EV9V test set (888 videos). Every entry of Figure~\ref{fig:perclass}(a) is the mean over the three runs of that model, whose run-to-run spread is given in Tables~\ref{tab:2d_baseline} and~\ref{tab:video_results} and is not repeated here; \textit{Sup.} is the number of test videos of that view and \textit{Macro} averages the nine views. In Figure~\ref{fig:perclass}(b) each row sums to 100 by construction, the colour scale covers only the off-diagonal, and the diagonal is shaded in grey because those nine values are the STFM recall column of the table beside them. Cell labels are rounded to the nearest whole percentage point, entries below 1\% are shown by colour alone, and a blank cell means that no test video of that view was assigned to that prediction in any of the three seeds; the matrix is averaged over those three STFM seeds.}
    \label{fig:perclass}
\end{figure*}

\subsection{Backbone Ablation}
Table~\ref{tab:ablation_backbone} compares ten CNN backbones across five architecture families under fixed temporal settings (hidden size 512, 2 LSTM layers). DenseNet-169 attains the best validation accuracy (95.36\%) and validation macro F1 (91.84\%), while the highest test accuracy (93.99\%) and test macro F1 (90.29\%) are obtained with ResNet-50. The strongest results are obtained by relatively compact architectures. Larger models such as ConvNeXt-Small (53.1M parameters) do not exhibit clear performance advantages, suggesting that increasing model capacity alone is insufficient to improve classification performance on EV9V. Similarly, lightweight MobileNetV3 variants (4.5M--6.6M parameters) still achieve over 92.9\% test accuracy, indicating that satisfactory performance can be achieved with substantially smaller models. The two backbones that rank above ResNet-18 on the validation set are less attractive on cost: DenseNet-169 requires roughly twice its FLOPs (17.04G vs.\ 8.63G), and EfficientNetV2-S uses 1.6$\times$ its parameters (23.8M vs.\ 14.7M) for a 0.12\% gain. Both validation margins are comparable to the seed-to-seed deviation of ResNet-18 ($\pm$0.27\%), whereas the lighter MobileNetV3 variants fall 1.82\% and 0.82\% behind it. Considering ResNet-18's competitive validation accuracy (95.00\%, within 0.36\% of the best) and its substantially smaller parameter count (14.7M against 23.8M for EfficientNetV2-S, at a comparable per-segment cost of 8.63G against 7.61G FLOPs), we adopt it as the default backbone in subsequent experiments.

\begin{table*}[!tb]
\centering
\caption{Backbone ablation. Temporal hidden size=512, layers=2. The best validation and test accuracies are in \textbf{bold}. All metrics are reported in percentage (\%).}
\label{tab:ablation_backbone}
\setlength{\tabcolsep}{11pt}
\begin{tabular}{llcccccc}
\toprule
\multicolumn{2}{c}{Model} & Params & FLOPs & Val Acc. & Val F1 & Test Acc. & Test F1 \\
\midrule
ResNet       & ResNet-18         & 14.7M & 8.63G  & $95.00 {\scriptscriptstyle \pm 0.27}$ & $90.94 {\scriptscriptstyle \pm 0.27}$ & $93.58 {\scriptscriptstyle \pm 0.00}$ & $89.67 {\scriptscriptstyle \pm 0.33}$ \\
             & ResNet-50         & 27.3M & 15.44G & $94.71 {\scriptscriptstyle \pm 0.47}$ & $90.36 {\scriptscriptstyle \pm 1.29}$ & $\mathbf{93.99} {\scriptscriptstyle \pm 0.68}$ & $90.29 {\scriptscriptstyle \pm 1.02}$ \\
\midrule
DenseNet     & DenseNet-121      & 10.6M & 15.97G & $94.83 {\scriptscriptstyle \pm 0.20}$ & $90.76 {\scriptscriptstyle \pm 0.29}$ & $93.51 {\scriptscriptstyle \pm 0.23}$ & $89.58 {\scriptscriptstyle \pm 0.22}$ \\
             & DenseNet-169      & 16.3M & 17.04G & $\mathbf{95.36} {\scriptscriptstyle \pm 0.41}$ & $91.84 {\scriptscriptstyle \pm 0.79}$ & $93.43 {\scriptscriptstyle \pm 0.36}$ & $89.33 {\scriptscriptstyle \pm 0.42}$ \\
\midrule
EfficientNet & EfficientNetV2-S & 23.8M & 7.61G  & $95.12 {\scriptscriptstyle \pm 0.27}$ & $90.98 {\scriptscriptstyle \pm 0.52}$ & $93.77 {\scriptscriptstyle \pm 0.64}$ & $89.90 {\scriptscriptstyle \pm 1.09}$ \\
             & EfficientNetV2-M & 56.5M & 13.23G & $94.53 {\scriptscriptstyle \pm 0.18}$ & $89.89 {\scriptscriptstyle \pm 0.46}$ & $93.69 {\scriptscriptstyle \pm 0.70}$ & $89.82 {\scriptscriptstyle \pm 0.97}$ \\
\midrule
ConvNeXt     & ConvNeXt-Tiny     & 31.4M & 15.33G & $94.24 {\scriptscriptstyle \pm 0.54}$ & $89.86 {\scriptscriptstyle \pm 1.08}$ & $93.06 {\scriptscriptstyle \pm 1.32}$ & $88.88 {\scriptscriptstyle \pm 1.71}$ \\
             & ConvNeXt-Small    & 53.1M & 23.80G & $94.12 {\scriptscriptstyle \pm 0.51}$ & $89.50 {\scriptscriptstyle \pm 1.13}$ & $93.28 {\scriptscriptstyle \pm 0.47}$ & $89.22 {\scriptscriptstyle \pm 1.14}$ \\
\midrule
MobileNetV3  & MobileNetV3-Small & 4.5M  & 0.36G  & $93.18 {\scriptscriptstyle \pm 0.57}$ & $87.88 {\scriptscriptstyle \pm 1.19}$ & $92.94 {\scriptscriptstyle \pm 0.33}$ & $88.68 {\scriptscriptstyle \pm 0.26}$ \\
             & MobileNetV3-Large & 6.6M  & 1.17G  & $94.18 {\scriptscriptstyle \pm 0.47}$ & $89.67 {\scriptscriptstyle \pm 0.95}$ & $93.24 {\scriptscriptstyle \pm 0.56}$ & $89.42 {\scriptscriptstyle \pm 0.83}$ \\
\bottomrule
\end{tabular}%
\end{table*}

\subsection{Temporal Modeling Ablation}
Table~\ref{tab:ablation_temporal_grid} presents a joint grid search over LSTM hidden size and number of layers, extended to include additional smaller configurations (hidden size 32 and 64). The best validation accuracy (95.30\%) is achieved with hidden size 128 and one layer, closely followed by hidden size 512 with one layer (95.00\%) and hidden size 512 with two layers (94.94\%). We retain hidden size 512 with two layers as the default: it lies within 0.36 percentage points of the best validation accuracy, whereas the best-on-validation configuration is clearly weaker on the test set (93.81\% against 94.29\%). Notably, competitive validation performance can be obtained across a wide range of temporal modeling capacities. Even compact configurations with hidden sizes of 32 or 64 achieve results comparable to those of much larger models. This suggests that the proposed fusion framework can effectively exploit temporal information without requiring large recurrent modules.

\begin{table*}[!tb]
\centering
\caption{Temporal grid: hidden size $\times$ layers. Backbone=ResNet-18. The best validation and test accuracies are in \textbf{bold}. All metrics are reported in percentage (\%).}
\label{tab:ablation_temporal_grid}
\setlength{\tabcolsep}{14pt}
\begin{tabular}{lcccccc}
\toprule
\multirow{2}{*}{Hidden Size} & \multicolumn{3}{c}{Val Acc.} & \multicolumn{3}{c}{Test Acc.} \\
\cmidrule{2-4} \cmidrule{5-7}
 & Layers=1 & Layers=2 & Layers=3 & Layers=1 & Layers=2 & Layers=3 \\
\midrule
32  & $94.77 {\scriptscriptstyle \pm 0.20}$ & $94.71 {\scriptscriptstyle \pm 0.47}$ & $94.59 {\scriptscriptstyle \pm 0.10}$ & $94.26 {\scriptscriptstyle \pm 0.34}$ & $94.14 {\scriptscriptstyle \pm 0.49}$ & $93.47 {\scriptscriptstyle \pm 0.39}$ \\
64  & $94.83 {\scriptscriptstyle \pm 0.10}$ & $94.77 {\scriptscriptstyle \pm 0.27}$ & $94.89 {\scriptscriptstyle \pm 0.00}$ & $93.66 {\scriptscriptstyle \pm 0.56}$ & $93.88 {\scriptscriptstyle \pm 0.75}$ & $94.14 {\scriptscriptstyle \pm 0.39}$ \\
128 & $\mathbf{95.30} {\scriptscriptstyle \pm 0.44}$ & $94.53 {\scriptscriptstyle \pm 0.18}$ & $94.83 {\scriptscriptstyle \pm 0.37}$ & $93.81 {\scriptscriptstyle \pm 0.92}$ & $94.14 {\scriptscriptstyle \pm 0.34}$ & $93.81 {\scriptscriptstyle \pm 0.52}$ \\
256 & $94.71 {\scriptscriptstyle \pm 0.53}$ & $94.83 {\scriptscriptstyle \pm 0.10}$ & $94.59 {\scriptscriptstyle \pm 0.57}$ & $94.14 {\scriptscriptstyle \pm 0.20}$ & $93.81 {\scriptscriptstyle \pm 0.39}$ & $94.14 {\scriptscriptstyle \pm 0.11}$ \\
512 & $95.00 {\scriptscriptstyle \pm 0.10}$ & $94.94 {\scriptscriptstyle \pm 0.54}$ & $94.12 {\scriptscriptstyle \pm 0.62}$ & $94.07 {\scriptscriptstyle \pm 0.28}$ & $94.29 {\scriptscriptstyle \pm 0.69}$ & $94.14 {\scriptscriptstyle \pm 0.23}$ \\
1024 & $94.53 {\scriptscriptstyle \pm 0.18}$ & $94.65 {\scriptscriptstyle \pm 0.20}$ & $94.59 {\scriptscriptstyle \pm 0.44}$ & $93.96 {\scriptscriptstyle \pm 0.17}$ & $\mathbf{94.44} {\scriptscriptstyle \pm 0.46}$ & $93.92 {\scriptscriptstyle \pm 0.52}$ \\
\bottomrule
\end{tabular}%
\end{table*}

\subsection{Re-EDL Hyperparameter Sensitivity}

The Re-EDL loss introduces a prior strength coefficient $\lambda_2$ (Equation~\ref{eq:uncertainty_u}) that controls the evidence strength. Table~\ref{tab:ablation_lamb2} evaluates $\lambda_2$ in the range of 0.2 to 2.0. The best validation accuracy is achieved at $\lambda_2 = $ $0.2$ (94.94\%), while the best test performance is obtained at $\lambda_2 = $ $1.0$ (94.44\% accuracy, 90.88\% macro F1). The default setting $\lambda_2 = 0.8$ achieves 94.18\% test accuracy and 90.51\% macro F1 in Table~\ref{tab:ablation_lamb2}, remaining within 0.26\% of the best test result. Overall, performance varies within the run-to-run variability of individual configurations across all settings, indicating that the model is relatively insensitive to the choice of prior strength within a practical interval. This stability suggests that the evidential learning behavior is primarily driven by the learned evidence rather than the exact weighting of the Dirichlet prior, allowing the model to maintain robust uncertainty estimation without requiring careful tuning of $\lambda_2$.
\begin{table*}[!tb]
\centering
\caption{$\lambda_2$ sensitivity analysis. Backbone=ResNet-18, temporal hidden size=512, layers=2, REEDL loss. All metrics are reported in percentage (\%).}
\label{tab:ablation_lamb2}
\setlength{\tabcolsep}{11pt}
\begin{tabular}{lcccccccc}
\toprule
\multicolumn{2}{c}{$\lambda_2$} & 0.2 & 0.4 & 0.6 & 0.8 & 1.0 & 1.5 & 2.0 \\
\midrule
\multirow{2}{*}{Val}  & Acc. & $\mathbf{94.94} {\scriptscriptstyle \pm 0.20}$ & $94.77 {\scriptscriptstyle \pm 0.20}$ & $94.89 {\scriptscriptstyle \pm 0.18}$ & $94.71 {\scriptscriptstyle \pm 0.31}$ & $94.59 {\scriptscriptstyle \pm 0.37}$ & $94.83 {\scriptscriptstyle \pm 0.27}$ & $94.77 {\scriptscriptstyle \pm 0.27}$ \\
                      & F1   & $90.90 {\scriptscriptstyle \pm 1.01}$ & $90.67 {\scriptscriptstyle \pm 0.19}$ & $90.86 {\scriptscriptstyle \pm 0.45}$ & $90.06 {\scriptscriptstyle \pm 0.56}$ & $90.02 {\scriptscriptstyle \pm 0.70}$ & $90.62 {\scriptscriptstyle \pm 0.80}$ & $90.47 {\scriptscriptstyle \pm 0.33}$ \\
\midrule
\multirow{2}{*}{Test} & Acc. & $93.88 {\scriptscriptstyle \pm 0.23}$ & $93.54 {\scriptscriptstyle \pm 0.53}$ & $93.24 {\scriptscriptstyle \pm 0.79}$ & $94.18 {\scriptscriptstyle \pm 0.46}$ & $\mathbf{94.44} {\scriptscriptstyle \pm 0.34}$ & $93.92 {\scriptscriptstyle \pm 0.34}$ & $93.77 {\scriptscriptstyle \pm 0.96}$ \\
                      & F1   & $89.99 {\scriptscriptstyle \pm 0.25}$ & $89.40 {\scriptscriptstyle \pm 0.76}$ & $89.22 {\scriptscriptstyle \pm 1.56}$ & $90.51 {\scriptscriptstyle \pm 0.78}$ & $90.88 {\scriptscriptstyle \pm 0.49}$ & $90.01 {\scriptscriptstyle \pm 0.61}$ & $90.03 {\scriptscriptstyle \pm 1.21}$ \\
\bottomrule
\end{tabular}%
\end{table*}

\subsection{Epsilon Sensitivity Analysis}

The selective sampling strategy adopts an $\epsilon$-greedy mechanism to balance exploitation of high-certainty segments and exploration of under-sampled regions. With probability $1-\epsilon$, the segment with the highest estimated certainty is selected, while with probability $\epsilon$, a random segment is sampled to prevent over-reliance on high-certainty observations. When $\epsilon = 1.0$, the strategy reduces to uniform random sampling, corresponding to removing the uncertainty-guided selection mechanism. Table~\ref{tab:ablation_epsilon} evaluates $\epsilon$ in the range of 0.1 to 1.0. The highest validation accuracy is achieved at $\epsilon = $ $0.8$ (95.00\%), while the best test performance is obtained at $\epsilon = $ $0.4$ (94.48\% test accuracy, 90.95\% macro F1). Within this range neither end performs best on the validation set: the accuracy at $\epsilon=1.0$, where the guidance is discarded entirely, stays below that of $\epsilon=0.8$, and the nearly deterministic selection at $\epsilon=0.1$ is the weakest of all, so guided selection and exploratory coverage have to be balanced rather than pushed to either extreme.

\begin{table*}[!tb]
\centering
\caption{$\epsilon$ sensitivity analysis. Backbone=ResNet-18, temporal hidden size=512, layers=2, REEDL loss with $\lambda_2=0.8$. All metrics are reported in percentage (\%).}
\label{tab:ablation_epsilon}
\setlength{\tabcolsep}{12pt}
\begin{tabular}{lccccccc}
\toprule
\multicolumn{2}{c}{$\epsilon$} & 0.1 & 0.2 & 0.4 & 0.6 & 0.8 & 1.0 \\
\midrule
\multirow{2}{*}{Val}  & Acc. & $94.42 {\scriptscriptstyle \pm 0.27}$ & $94.65 {\scriptscriptstyle \pm 0.54}$ & $94.77 {\scriptscriptstyle \pm 0.37}$ & $94.71 {\scriptscriptstyle \pm 0.53}$ & $\mathbf{95.00} {\scriptscriptstyle \pm 0.10}$ & $94.89 {\scriptscriptstyle \pm 0.31}$ \\
                      & F1   & $89.51 {\scriptscriptstyle \pm 0.79}$ & $90.18 {\scriptscriptstyle \pm 1.13}$ & $90.15 {\scriptscriptstyle \pm 0.89}$ & $89.91 {\scriptscriptstyle \pm 1.16}$ & $90.84 {\scriptscriptstyle \pm 0.28}$ & $90.85 {\scriptscriptstyle \pm 0.47}$ \\
\midrule
\multirow{2}{*}{Test} & Acc. & $93.77 {\scriptscriptstyle \pm 0.33}$ & $93.99 {\scriptscriptstyle \pm 0.51}$ & $\mathbf{94.48} {\scriptscriptstyle \pm 0.49}$ & $94.37 {\scriptscriptstyle \pm 0.11}$ & $93.99 {\scriptscriptstyle \pm 1.02}$ & $93.77 {\scriptscriptstyle \pm 0.17}$ \\
                      & F1   & $89.96 {\scriptscriptstyle \pm 0.64}$ & $90.18 {\scriptscriptstyle \pm 0.68}$ & $90.95 {\scriptscriptstyle \pm 0.67}$ & $90.79 {\scriptscriptstyle \pm 0.19}$ & $90.24 {\scriptscriptstyle \pm 1.74}$ & $89.82 {\scriptscriptstyle \pm 0.53}$ \\
\bottomrule
\end{tabular}%

\end{table*}

\subsection{\texorpdfstring{Data Scaling Analysis}{Data Scaling Analysis}}

To assess how the amount of training data affects the proposed model, we retrain STFM with its default configuration on subsets randomly drawn from the EV9V training split, while the validation and test splits are kept fixed. Table~\ref{tab:data_scaling} reports the four settings, from 905 videos (24.6\% of the training split) to the full 3,683, each trained with three random seeds; the full-split column is the default configuration reported in Table~\ref{tab:ablation_epsilon}.

Validation accuracy improves monotonically with the amount of training data, from 93.83\% to 95.00\%, and validation macro F1 from 88.50\% to 90.84\%. The improvement on the test set is weaker: from the smallest subset to the full split, test accuracy rises from 93.73\% to 93.99\% and macro F1 from 89.65\% to 90.24\%, with a dip to 93.32\% at 72.2\%. The seed-to-seed standard deviation of individual settings reaches $\pm$1.02\% in accuracy and $\pm$1.74\% in macro F1, of the same order as these differences, so the test-set trend is a mild improvement rather than a pronounced scaling effect. STFM nevertheless reaches close to its full-data test accuracy with about a quarter of the training videos. The full training split is retained as the default, as it gives the highest validation and test accuracy and macro F1 of the four settings.
\begin{table*}[!tb]
\centering
\caption{Data scaling analysis. STFM with its default configuration is retrained on subsets of the EV9V training split, with the validation and test splits fixed. Backbone=ResNet-18, temporal hidden size=512, layers=2, REEDL loss with $\lambda_2=0.8$ and $\epsilon=0.8$. All metrics are reported in percentage (\%) and averaged over three random seeds. The best validation and test accuracies are shown in \textbf{bold}.}
\label{tab:data_scaling}
\setlength{\tabcolsep}{12pt}
\begin{tabular}{lccccc}
\toprule
\multicolumn{2}{c}{Training split} & 24.6\% & 48.1\% & 72.2\% & 100\% \\
\multicolumn{2}{c}{Training videos} & 905 & 1,772 & 2,659 & 3,683 \\
\midrule
\multirow{2}{*}{Val}  & Acc. & $93.83 {\scriptscriptstyle \pm 0.31}$ & $93.94 {\scriptscriptstyle \pm 0.27}$ & $94.06 {\scriptscriptstyle \pm 0.20}$ & $\mathbf{95.00} {\scriptscriptstyle \pm 0.10}$ \\
                      & F1   & $88.50 {\scriptscriptstyle \pm 0.11}$ & $88.57 {\scriptscriptstyle \pm 0.72}$ & $89.02 {\scriptscriptstyle \pm 0.61}$ & $90.84 {\scriptscriptstyle \pm 0.28}$ \\
\midrule
\multirow{2}{*}{Test} & Acc. & $93.73 {\scriptscriptstyle \pm 0.62}$ & $93.84 {\scriptscriptstyle \pm 0.23}$ & $93.32 {\scriptscriptstyle \pm 0.53}$ & $\mathbf{93.99} {\scriptscriptstyle \pm 1.02}$ \\
                      & F1   & $89.65 {\scriptscriptstyle \pm 0.97}$ & $89.79 {\scriptscriptstyle \pm 0.21}$ & $88.96 {\scriptscriptstyle \pm 0.88}$ & $90.24 {\scriptscriptstyle \pm 1.74}$ \\
\bottomrule
\end{tabular}%

\end{table*}

\section{Interpretability Analysis}
\label{sec:interpretability}

\subsection{Error Case Analysis}

\begin{figure*}[!tb]
\centering
\scriptsize
\setlength{\tabcolsep}{1.5pt}

\begin{tabular}{c c c c c c}
\toprule
\shortstack{Confusion\\Pair} &
\shortstack{Reference\\(Ground Truth)} &
\multicolumn{3}{c}{\shortstack{Misclassified\\Samples}} &
\shortstack{Reference\\(Predicted Class)} \\
\midrule

\scriptsize\shortstack{PMASA\\$\rightarrow$ PMPALA}
&
\includegraphics[width=0.155\textwidth]{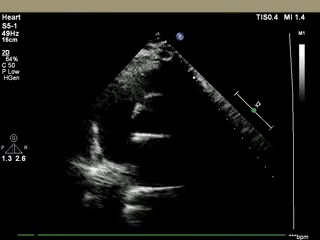}
&
\includegraphics[width=0.155\textwidth]{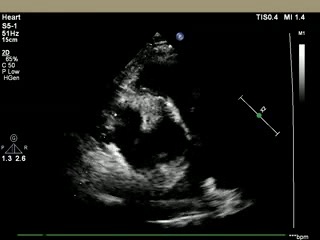}
&
\includegraphics[width=0.155\textwidth]{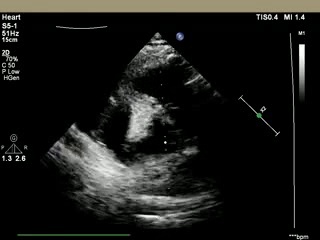}
&
\includegraphics[width=0.155\textwidth]{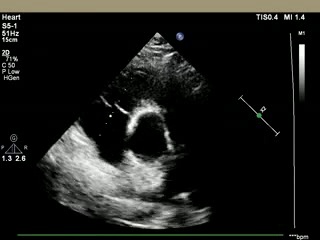}
&
\includegraphics[width=0.155\textwidth]{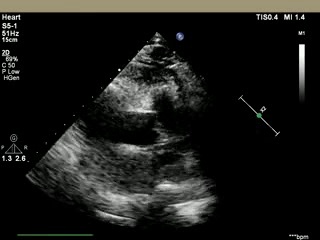}
\\[8pt]

\scriptsize\shortstack{A4C\\$\rightarrow$ A5C}
&
\includegraphics[width=0.155\textwidth]{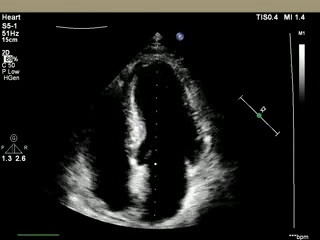}
&
\includegraphics[width=0.155\textwidth]{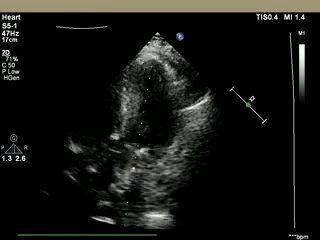}
&
\includegraphics[width=0.155\textwidth]{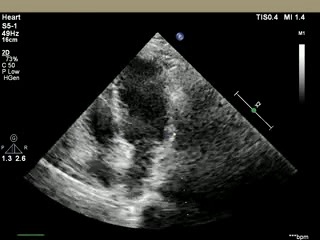}
&
\includegraphics[width=0.155\textwidth]{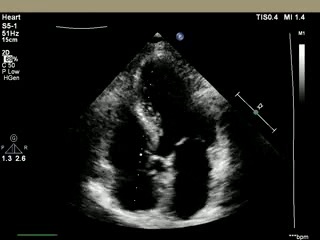}
&
\includegraphics[width=0.155\textwidth]{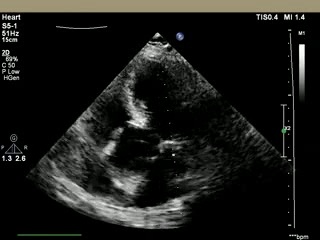}
\\[8pt]

\scriptsize\shortstack{PMPALA\\$\rightarrow$ PMASA}
&
\includegraphics[width=0.155\textwidth]{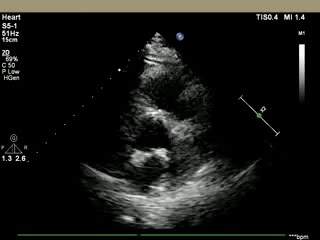}
&
\includegraphics[width=0.155\textwidth]{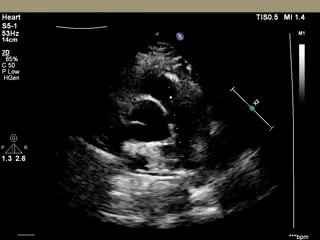}
&
\includegraphics[width=0.155\textwidth]{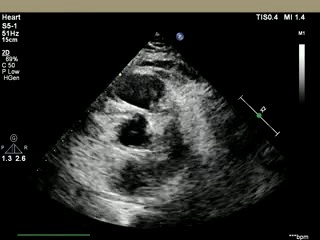}
&
\includegraphics[width=0.155\textwidth]{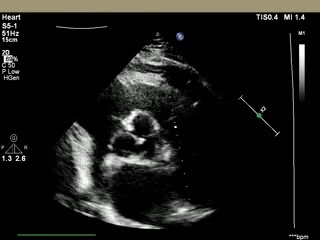}
&
\includegraphics[width=0.155\textwidth]{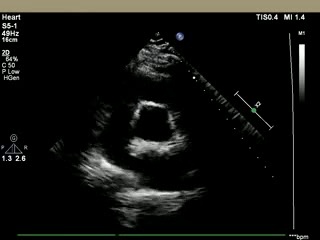}
\\[8pt]

\scriptsize\shortstack{PASA\\$\rightarrow$ PPMLSA}
&
\includegraphics[width=0.155\textwidth]{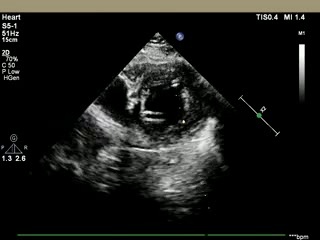}
&
\includegraphics[width=0.155\textwidth]{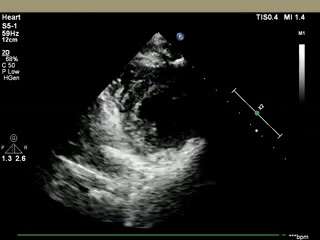}
&
\includegraphics[width=0.155\textwidth]{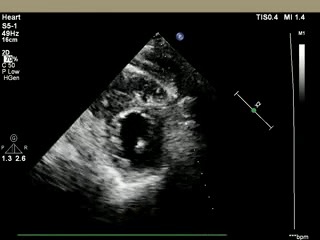}
&
\includegraphics[width=0.155\textwidth]{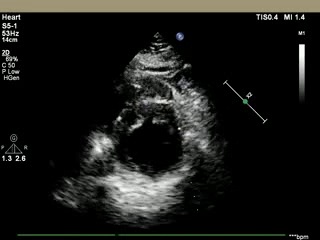}
&
\includegraphics[width=0.155\textwidth]{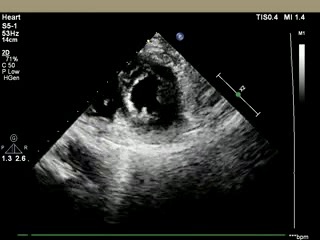}
\\[8pt]

\scriptsize\shortstack{PPMLSA\\$\rightarrow$ PASA}
&
\includegraphics[width=0.155\textwidth]{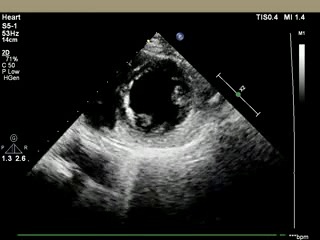}
&
\includegraphics[width=0.155\textwidth]{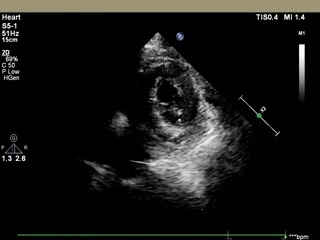}
&
\includegraphics[width=0.155\textwidth]{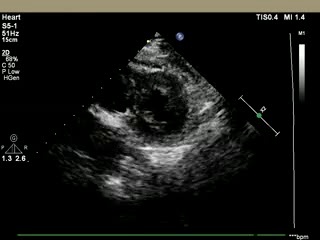}
&
\includegraphics[width=0.155\textwidth]{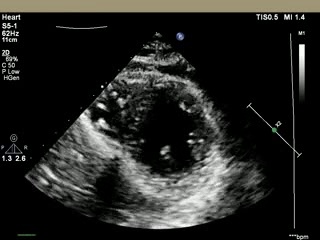}
&
\includegraphics[width=0.155\textwidth]{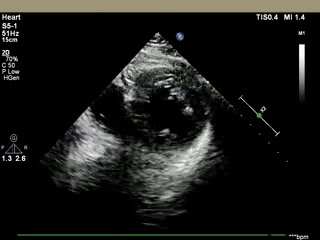}
\\

\bottomrule
\end{tabular}

\caption{
Representative examples from the five most frequent confusion pairs on the EV9V test set. Each row corresponds to a confusion direction (ground-truth $\rightarrow$ predicted class). The first column of images shows a correctly classified example from the ground-truth class, the last column shows a correctly classified example from the predicted class, and the three middle columns present representative misclassified samples.
}
\label{fig:error_cases}

\end{figure*}
Figure~\ref{fig:perclass} compares the per-class precision, recall and F1 score of RegNet, UniFormerV2, and the proposed STFM on the EV9V test set, together with the confusion matrix of STFM, highlighting how temporal information mitigates the limitations of frame-level classification.

Two observations can be made. First, classification performance is not strongly correlated with class frequency. For example, PLHLA (the largest class with 275 test samples) achieves perfect recall across all models, while SC4C (one of the smallest classes with 35 samples) also obtains over 97\% recall. This indicates that class frequency is not the only factor: SC4C, one of the smallest views (35 test videos), reaches the third-highest F1 (96.7), while the four lowest F1 scores all belong to the five smallest views and the rank correlation between view frequency and per-class F1 is only moderate (Spearman $\rho=0.48$, $n=9$, $p\approx0.19$); the intrinsic anatomical discriminability of each view therefore plays a more critical role in classification difficulty. Second, certain anatomically similar views remain consistently challenging across models, while benefiting differently from temporal modeling. In particular, the three models handle A5C differently: STFM attains the highest recall (86.7\%, against 85.0\% for UniFormerV2 and 71.7\% for RegNet) but the lowest precision (83.8\%, against 90.3\% and 93.6\%), so it recovers more of the A5C videos at the cost of predicting that view more freely, whereas UniFormerV2 obtains the best F1 on this view (87.5\%, against 85.1\% for STFM and 80.9\% for RegNet); the STFM--RegNet difference of 4.2 points is smaller than the seed-to-seed spread of this view ($\pm$5.1 and $\pm$5.6). The confusion matrix of Figure~\ref{fig:perclass}(b) localises what remains: 12\% of the A5C videos are labelled A4C, against 2\% of the A4C videos labelled A5C, and the five largest off-diagonal entries, 20\% for PASA to PPMLSA, 16\% for PMASA to PMPALA, 12\% for A5C to A4C, 11\% for PPMLSA to PASA and 9\% for PMPALA to PMASA, all fall between the pairs of views examined in Figure~\ref{fig:error_cases}.Similar trends are observed for PASA and PPMLSA, where video-based models generally achieve higher or comparable recall relative to frame-level approaches. The 95\% Wilson intervals of recall span 14--24 percentage points for the five views with fewer than 60 test videos, against 1.4 and 4.5 percentage points for the two largest views, so a difference of a few videos on a small view does not separate two models and the macro average is the more stable reading of Figure~\ref{fig:perclass}(a).

 Figure~\ref{fig:error_cases} visualizes the five most frequent confusion pairs, where misclassifications predominantly occur between anatomically adjacent or visually similar views. Each row shows five images: a correctly classified frame of the ground-truth class (first column), three misclassified frames (middle columns), and a correctly classified frame of the predicted class (last column) for reference. The dominant confusion pairs can be categorized into three types. The first type involves anatomically adjacent planes that share substantial local structure. For example, PMASA and PMPALA are both parasternal short-axis views distinguished primarily by the presence of the pulmonary artery bifurcation, which occupies only a small image region. Similarly, PASA and PPMLSA differ mainly by the visibility of papillary muscles at different left ventricular levels, and misclassification occurs when these structures are not clearly resolved due to poor image quality. The second type involves views where one is a structural subset of the other. A5C is essentially A4C with an additional aortic root and LVOT; when this small discriminative feature is incompletely visualized, the model confuses the two views in either direction. The third type arises from transitional views---non-standard frames captured during probe adjustment, respiration, or cardiac motion---which naturally lie between two standard planes and can be ambiguous even for human annotation. Suboptimal image quality, including blurring and poor acoustic windows, further exacerbates these ambiguities. 

\subsection{Uncertainty Visualization}

\begin{figure*}[!tb]
    \centering
    \begin{subfigure}{\textwidth}
        \centering
        \includegraphics[width=\textwidth]{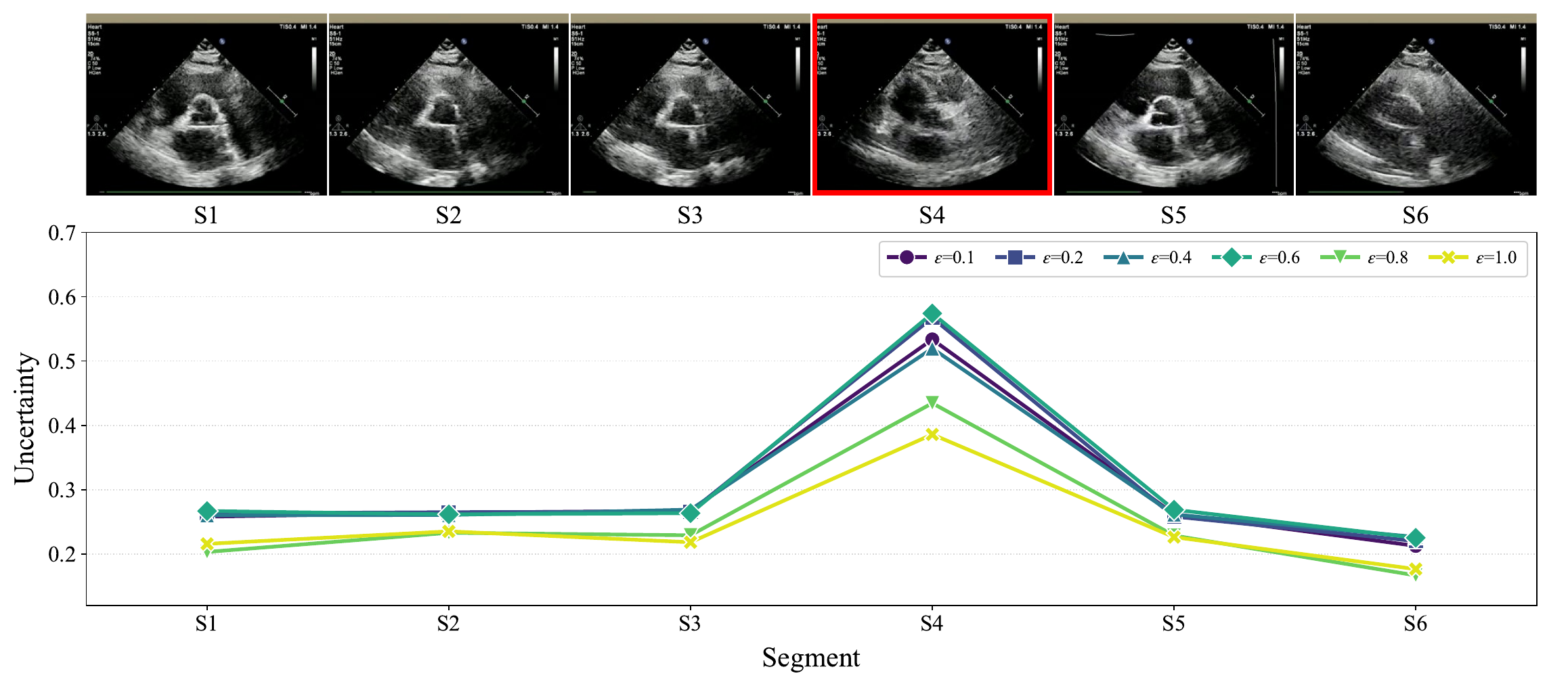}
        \caption{Segment-level uncertainty trajectories for a representative validation video (true label PMPALA) under all six $\epsilon$ configurations (three-seed means). The frame highlighted in red marks segment S4, at which every configuration attains its maximum.}
        \label{fig:uncertainty_a}
    \end{subfigure}
    \vspace{8pt}
    \begin{subfigure}[t]{0.48\textwidth}
        \centering
        \includegraphics[width=\linewidth]{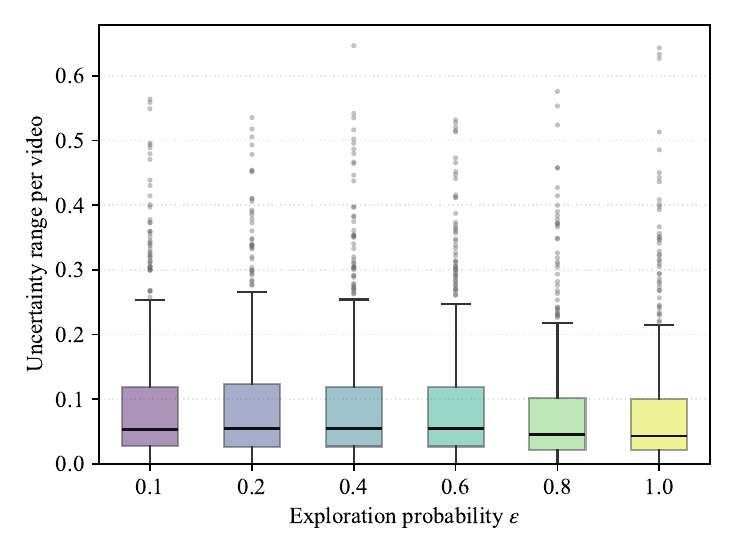}
        \caption{Per-video uncertainty range (max-min across segments) on the validation set, for the six $\epsilon$ configurations.}
        \label{fig:uncertainty_b}
    \end{subfigure}
    \hfill
    \begin{subfigure}[t]{0.48\textwidth}
        \centering
        \includegraphics[width=\linewidth]{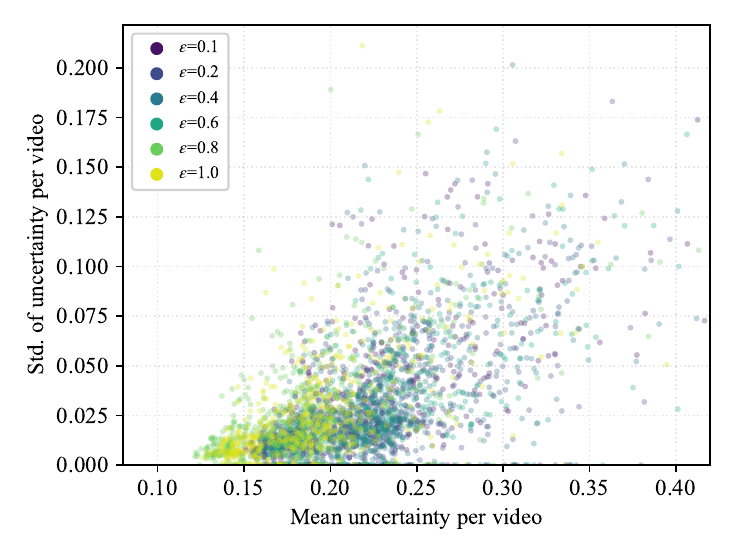}
        \caption{Per-video mean vs.\ standard deviation of uncertainty on the validation set, for the six $\epsilon$ configurations.}
        \label{fig:uncertainty_c}
    \end{subfigure}
    \caption{Uncertainty analysis of STFM under different $\epsilon$ configurations.}
    \label{fig:uncertainty}
\end{figure*}

We further analyze the evidential uncertainty produced by the Re-EDL head for the six exploration probabilities $\epsilon\in\{0.1,0.2,0.4,0.6,0.8,1.0\}$ examined in Table~\ref{tab:ablation_epsilon}, each trained with three random seeds. Figure~\ref{fig:uncertainty}(a) shows the segment-level uncertainty trajectories of these configurations on a representative validation video whose true label is PMPALA. The video is divided into 10 equally spaced segments: the frames at the centers of the first six are displayed along the top, while the bottom section plots the corresponding uncertainty values.

Across all six configurations, uncertainty reaches its maximum at segment S4, where the three-seed mean ranges from 0.39 ($\epsilon=1.0$) to 0.57 ($\epsilon=0.6$), while every other segment of the video stays below 0.40. The frame at S4 is a transitional view in which the cardiac anatomy is poorly defined. The peak coincides with the error: at this segment the highest-probability class is PMASA in all 18 runs (mean probability 0.671, against 0.144 for the true class), the same wrong label that the video receives. The agreement of the peak location across seeds and exploration probabilities indicates that the uncertainty signal tracks classification errors rather than a particular training configuration.

We next examine this behavior at the dataset level. Figures~\ref{fig:uncertainty}(b) and (c) summarize per-video uncertainty statistics over the 567 validation videos for each configuration, computed from the three-seed mean segment curves. Figure~\ref{fig:uncertainty}(b) shows the per-video uncertainty range (maximum minus minimum uncertainty across segments). The configurations with $\epsilon\le0.6$ exhibit similar distributions, with median ranges of 0.053--0.055, whereas $\epsilon=0.8$ and $\epsilon=1.0$ are lower (0.046 and 0.043). Figure~\ref{fig:uncertainty}(c) plots the per-video standard deviation of uncertainty against its mean and shows the same pattern.

Taken together, the two panels carry the segment-level picture to the whole validation set: the uncertainty estimates of a video are more spread out across its segments when the selection is guided by uncertainty than when segments are chosen at random, whose median per-video range is the narrowest of the six configurations (0.043, against 0.053--0.055 for $\epsilon\le0.6$). Selective sampling therefore helps the model retain sensitivity to atypical segments, since the uncertainty estimates of one video stay differentiated across its segments.

\subsection{\texorpdfstring{Confidence Calibration}{Confidence Calibration}}
\label{sec:calibration}

\begin{figure*}[!tb]
\centering
\begin{subfigure}[t]{0.49\textwidth}
        \centering
        \includegraphics[width=\linewidth]{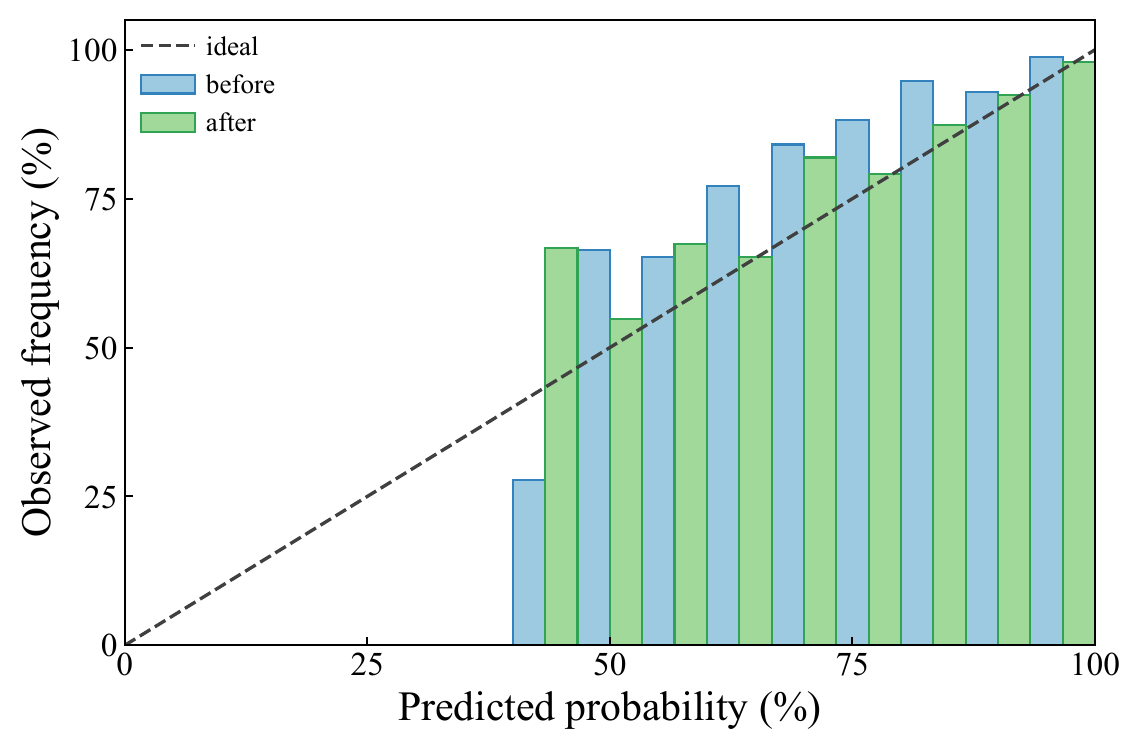}
        \caption{STFM, before and after calibration.}
        \label{fig:reliability_a}
    \end{subfigure}
    \hfill
    \begin{subfigure}[t]{0.49\textwidth}
        \centering
        \includegraphics[width=\linewidth]{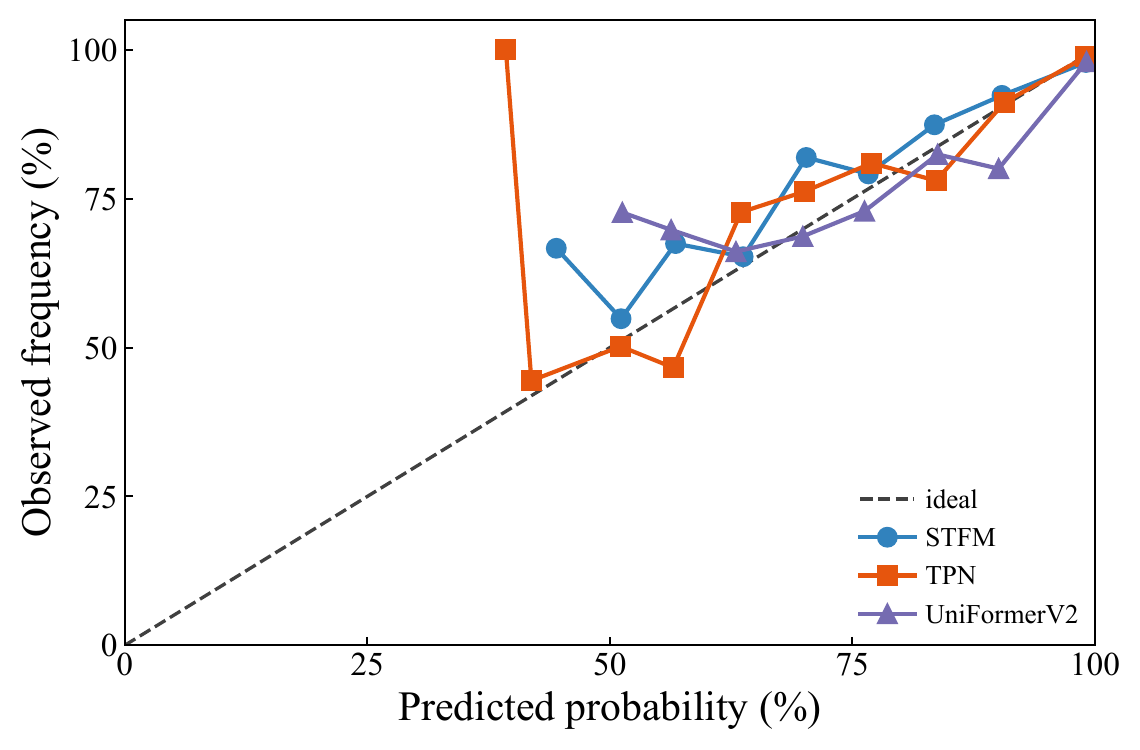}
        \caption{Comparison after calibration.}
        \label{fig:reliability_b}
    \end{subfigure}
    \caption{Reliability diagrams on the test set, with the 15 equal-width bins and the frozen temperature of Table~\ref{tab:calibration}. The dashed line is ideal and bars above it mark under-confidence; the STFM bars average its three seeds and the TPN and UniFormerV2 bars their three runs within each bin, so they need not reproduce the averaged ECE of the table. Bins below 46\% predicted probability hold under 0.4\% of the videos.}
\label{fig:reliability}
\end{figure*}

The calibration of the video-level probabilities was measured by fitting one temperature per model on the validation set and keeping it frozen for the test set, which rescales confidence without touching any prediction~\citep{guo2017calibration}. Table~\ref{tab:calibration} gives the expected and adaptive calibration errors (ECE and ACE, 15 bins)~\citep{naeini2015obtaining} before and after that step, and the multiclass Brier score~\citep{glenn1950verification} and NLL after it, for STFM and the two most accurate baselines; Figure~\ref{fig:reliability} shows the reliability diagrams.

The shortfall is in the stated confidence rather than in the predictions: STFM assigns a mean probability of 0.899 to the class it selects, against 93.99\% accuracy, whereas the two baselines state 0.966 and 0.949 against their 94.41\% and 94.22\% accuracies, and the temperatures fitted to them lie above 1 while STFM's is 0.72. The direction comes from the evidential read-out itself, a Dirichlet mean that carries $\lambda_2$ units of prior on every class besides the pooled evidence and is therefore pulled toward the uniform value, a shift that leaves the ordering of the nine classes untouched. Rescaling recovers part of it: STFM ends at 2.64\% ECE, still the highest of the three models, although by a narrow margin (1.63\% for TPN and 2.58\% for UniFormerV2), and the ACE of TPN even increases slightly (1.45\% to 1.51\%).

\begin{table*}[!tb]
\centering
\caption{Calibration of the video-level probabilities on the 888 test videos. $T$ is fitted on the validation set and frozen, so predictions are unchanged; Brier and NLL are reported after calibration. ECE and ACE use 15 equal-width and 15 equal-mass bins and are in percentage (\%). STFM is averaged over its three seeds (ECE spread $\pm$1.45 before, $\pm$0.23 after); each baseline is averaged over its three runs, with $T$ fitted on the validation dump of the first run and frozen.}
\label{tab:calibration}
\setlength{\tabcolsep}{12pt}
\begin{tabular}{lcccccccc}
\toprule
 & & & \multicolumn{2}{c}{ECE} & \multicolumn{2}{c}{ACE} & \multicolumn{2}{c}{Calibrated} \\
\cmidrule(lr){4-5} \cmidrule(lr){6-7} \cmidrule(lr){8-9}
Model & Acc. & $T$ & before & after & before & after & Brier & NLL \\
\midrule
STFM (Ours) & 93.99 & 0.72 & 4.17 & 2.64 & 4.32 & 2.67 & 0.1001 & 0.2119 \\
UniFormerV2 & 94.41 & 1.20 & 3.21 & 2.58 & 2.91 & 2.40 & 0.0931 & 0.1793 \\
TPN & 94.22 & 1.09 & 1.68 & 1.63 & 1.45 & 1.51 & 0.0864 & 0.1718 \\
\bottomrule
\end{tabular}%
\end{table*}

\section{Discussion}
\label{sec:discussion}

The benefits of STFM follow from the two mechanisms it is built on. A spatial stream describes the appearance of individual frames while a temporal stream captures the motion between consecutive frames, so anatomy and dynamics both enter the decision, and the frame-level evidence is pooled into one video-level distribution, which lets the observations that are best supported carry the final label. Selective sampling keeps that evidence drawn from representative segments while still covering the whole examination. The resulting model stays comparable to the strongest compared video architectures in accuracy while using a fraction of their parameters and computation, and it supplies an uncertainty that marks ambiguous segments without adding any cost at inference.

Several limitations should be noted. All recordings come from a single centre and follow one acquisition routine, and videos obtained with other devices or at other institutions were not available to us, so the present results say little about other imaging chains. The screening that makes the annotations trustworthy also narrows the cohort: examinations with poor acoustic windows or unusual probe positions were excluded, which leaves the harder real-world cases under-represented. The nine standard views are treated as a closed set, so a study in which a required view is absent, or in which two standard views share one clip, does not fit the present label scheme. The evidential uncertainty is, finally, only an indication that a prediction may deserve a second look; whether acting upon it changes what an examination workflow produces has not been examined.

These boundaries point at what comes next. Testing the same framework on multi-centre recordings acquired with different equipment, and moving from a single view label to the presence of views and the timing of events within an examination, would bring the task closer to the way results are reported in practice. The uncertainty signal is a natural basis for rejection, where the model answers only when its evidence is sufficient and passes ambiguous clips to a physician. Stronger backbones and self-supervised pre-training remain to be tried, since how much of the remaining margin belongs to the encoder is still open. The practical value of the method will ultimately be decided by how well it holds up under uneven acquisition quality.

\section{Conclusion}
\label{sec:conclusion}

In this paper, we introduced EV9V, a publicly accessible echocardiographic video dataset comprising 5,138 videos, 910,579 frames, and 9 standard views. To our knowledge, EV9V is the largest public dataset of its kind, featuring comprehensive view coverage, rigorous three-stage physician quality control, and substantial real-world variability. Using EV9V, we established a systematic benchmark for video-level echocardiographic view classification, evaluating a diverse range of architectures including 2D CNNs, 2D CNNs with temporal aggregation, RNN-based models, 3D CNNs, and video transformers under a unified protocol.

We proposed STFM, an efficient dual-stream CNN--LSTM architecture that jointly models spatial anatomical structures and temporal cardiac dynamics. By combining STFM with Reliable Evidential Deep Learning and an uncertainty-guided segment selection strategy, our method learns to preferentially sample representative observations during training and performs evidence-based fusion during inference. Extensive experiments demonstrated that STFM achieves competitive accuracy (93.99\% test accuracy) with substantially fewer parameters (14.75M) and lower computational cost (43.13G FLOPs) compared to state-of-the-art video architectures. 
Together, the dataset, the evaluation protocol and the model are intended as a reproducible starting point for further work on echocardiographic video analysis.

\section*{Acknowlegements}
This study was supported by grants from the National Natural Science
Foundation of China (No. 62376169), Scientific Research Project of Chengdu
Medical (No. 2023077, 2023279, and CYZYB25-20), Scientific Research Project of Chengdu Science and Technol-ogy Bureau (No. 2022-YF05-01414-SN), and Health Commission of Sichuan Province Medical Science and Technology Program (No. 26WXQNMP33).

\section*{Declaration of generative AI and AI-assisted technologies}
Statement: During the preparation of this work, the author(s) used ChatGPT for language 
polishing. After using this tool, the author(s) reviewed and edited the content as needed 
and take(s) full responsibility for the content of the published article.

\bibliographystyle{cas-model2-names}
\bibliography{ref}

\end{document}